\documentclass{ieeeaccess}
\usepackage{cite}
\usepackage{amsmath,amssymb,amsfonts}
\usepackage{algorithmic}
\usepackage{graphicx}
\usepackage{textcomp}
\usepackage{bm}
\usepackage{booktabs}
\usepackage{multirow}
\usepackage{caption}
\usepackage{subcaption}
\usepackage{hyperref}
\usepackage{tabularx}
\usepackage{multirow}
\makeatletter
\AtBeginDocument{\DeclareMathVersion{bold}
\SetSymbolFont{operators}{bold}{T1}{times}{b}{n}
\SetSymbolFont{NewLetters}{bold}{T1}{times}{b}{it}
\SetMathAlphabet{\mathrm}{bold}{T1}{times}{b}{n}
\SetMathAlphabet{\mathit}{bold}{T1}{times}{b}{it}
\SetMathAlphabet{\mathbf}{bold}{T1}{times}{b}{n}
\SetMathAlphabet{\mathtt}{bold}{OT1}{pcr}{b}{n}
\SetSymbolFont{symbols}{bold}{OMS}{cmsy}{b}{n}
\renewcommand\boldmath{\@nomath\boldmath\mathversion{bold}}}
\makeatother

\def\BibTeX{{\rm B\kern-.05em{\sc i\kern-.025em b}\kern-.08em
    T\kern-.1667em\lower.7ex\hbox{E}\kern-.125emX}}

\begin{document}
\history{Received 7 July 2026, accepted 17 August 2026, date of publication 28 August 2026, date of current version 9 September 2026.}
\doi{10.1109/ACCESS.2026.3728110}

\title{Same Methods, Different Rankings: Trainable Depth as an Evaluation Variable in Continual Learning}
\author{\uppercase{Paul-Tiberiu Iordache}\authorrefmark{1},
\uppercase{Elena Burceanu}\authorrefmark{1,2}, AND \uppercase{Mihai Dascalu}\authorrefmark{2,3}}

\address[1]{Bitdefender, Romania (e-mail: tiiordache@bitdefender.com, eburceanu@bitdefender.com)}
\address[2]{National University of Science and Technology POLITEHNICA Bucharest, Romania (e-mail: mihai.dascalu@upb.ro)}
\address[3]{Academy of Romanian Scientists, Romania}

\markboth
{Iordache \headeretal: Same Methods, Different Rankings: Trainable Depth as an Evaluation Variable in Continual Learning}
{Iordache \headeretal: Same Methods, Different Rankings: Trainable Depth as an Evaluation Variable in Continual Learning}

\corresp{Corresponding author: Elena Burceanu (e-mail: eburceanu@bitdefender.com).}

\begin{abstract}
Continual learning (CL) examines how models learn a sequence of tasks while retaining previously learned knowledge. Despite substantial progress in benchmarking CL methods, comparative evaluations typically keep the fine-tuning regime fixed. In this paper, we argue that the fine-tuning regime, defined by the trainable parameter subspace, is itself a key evaluation variable. We formalize adaptation regimes as projected optimization over fixed trainable subspaces, showing that changing the trainable depth alters the effective update signal through which both current task fitting and knowledge preservation operate. This analysis motivates the hypothesis that method comparisons need not be invariant across regimes. We test this hypothesis in task incremental CL while considering 5 trainable depth regimes and 5 standard methods: online EWC, LwF, SI, GEM, and DER. We find that the relative ranking of methods is not consistently preserved across regimes when evaluating across 5 benchmark datasets, namely MNIST, Fashion MNIST, KMNIST, QMNIST, and CIFAR-100, and across 11 task orders per dataset. We further show that deeper adaptation regimes are associated with larger update magnitudes, higher forgetting, and a stronger relationship between the two. These results show that comparative conclusions in CL can depend strongly on the chosen fine-tuning regime, motivating regime-aware evaluation protocols that treat trainable depth as an explicit experimental factor.
\end{abstract}


\titlepgskip=-21pt

\maketitle

\section{Introduction}

Building models that can integrate new information over time without erasing previously acquired knowledge is a long-standing challenge, central to deploying machine learning systems in dynamic, real-world environments \cite{Parisi2019ContinualLL, DeLange2022ACL}. At the core of CL lies the plasticity-stability trade-off: a model must remain plastic enough to learn new tasks while stable enough to avoid overwriting previously acquired knowledge, a phenomenon known as catastrophic forgetting \cite{McCloskey1989CatastrophicII, Kirkpatrick2017OvercomingCF}. A rich ecosystem of methods has emerged to address this, spanning regularization-based approaches \cite{Kirkpatrick2017OvercomingCF, zenke2017continuallearningsynapticintelligence}, distillation-based methods \cite{li2017learningforgetting}, and gradient-projection strategies \cite{lopezpaz2022gradientepisodicmemorycontinual}, each making different assumptions about the nature of task interference. Yet despite this progress, these methods are almost universally evaluated under a single fixed training setup \cite{Farquhar2018TowardsRC, Vandeven2019ThreeSF}, making it difficult to draw conclusions that generalize across different deployment conditions.

A key factor that is routinely left uncontrolled is the adaptation regime, that is, which parameters are kept trainable throughout the CL sequence. This choice directly shapes the optimization geometry available to the model \cite{Goodfellow2015AnEI} and fundamentally alters how the plasticity-stability trade-off manifests for each method. 

Regularization-based methods such as online Elastic Weight Consolidation (online EWC) and Synaptic Intelligence (SI) \cite{schwarz2018progresscompressscalable, zenke2017continuallearningsynapticintelligence}, distillation-based approaches like Learning without Forgetting (LwF) \cite{li2017learningforgetting}, gradient-projection methods such as Gradient Episodic Memory (GEM) \cite{lopezpaz2022gradientepisodicmemorycontinual}, and and dynamic architecture methods such as Dynamically Expandable Representation (DER) \cite{yan2021derdynamicallyexpandablerepresentation} each interact differently with the trainable subspace. Notably, none of these methods is normative about which subset of parameters should remain trainable, nor do they provide explicit guidance on this choice. As a result, there is no reason to expect their relative effectiveness to remain constant as the trainable depth varies \cite{Kumar2022FineTuningCD}. Yet this interaction is almost never studied explicitly, and method comparisons are routinely drawn under a single fixed regime without acknowledging the dependence this introduces \cite{Biesialska2020ContinualLL}. We study this interaction by treating fine-tuning regimes as constrained optimization over parameter subspaces, enabling a systematic analysis of how CL method rankings depend on trainable depth and motivating the need for more controlled, regime-aware evaluation protocols.

The choice of trainable depth fundamentally reshapes the geometry and capacity of the optimization space in which learning unfolds. When updates are restricted to a limited subset of layers, the model is compelled to encode all incoming task-specific information within a constrained parameter subspace, effectively compressing both adaptation and retention mechanisms into a narrower channel. This concentration of representational burden amplifies competition between tasks, as newly acquired knowledge must coexist with, and often overwrite, previously stored information within the same limited set of parameters. In contrast, full-network fine-tuning allows updates to be distributed across a higher-dimensional space, enabling the model to allocate distinct regions of the parameter space to different tasks and thereby mitigate interference. As a result, the degree of trainable depth directly governs not only the model’s expressive flexibility but also the dynamics of interference and retention, making the performance of any continual learning algorithm inherently dependent on the structure and dimensionality of the underlying trainable subspace.

Standard benchmarks typically evaluate methods using full-parameter updates across the entire architecture \cite{Lomonaco2017CORe50AN, Hsu2018ReevaluatingCA, Vandeven2019ThreeSF}, which may present an incomplete picture of algorithmic performance. If method rankings shift when the trainable subspace is restricted, then the perceived superiority of an algorithm might be conditional on the training regime rather than its intrinsic logic. This suggests that many comparative conclusions in the literature are regime-specific, necessitating more controlled, regime-aware protocols to ensure the robustness of continual learning methods.

Our main contributions are as follows:

\begin{itemize}

\item \textbf{A subspace view of adaptation regimes.} We formalize fine-tuning regimes in continual learning as constrained optimization over fixed trainable subspaces. Under this view, changing the regime changes not only the number of trainable parameters but also the feasible subspace in which optimization can occur throughout the task sequence. This makes explicit why the adaptation regime affects the effective update signal seen by the learner, and why it should be treated as a meaningful evaluation variable rather than a background implementation detail.

\item \textbf{A systematic empirical study across methods, datasets, and depths.} We compare 5 standard continual learning methods, online EWC, LwF, SI, GEM, and DER, across 5 trainable depth regimes, from training only the last residual block to full fine-tuning. We evaluate these methods on 5 benchmark datasets, namely MNIST, Fashion MNIST, KMNIST, QMNIST, and CIFAR-100, and across 11 sampled task orders for each dataset. Our results show that the relative ranking of methods can change substantially across regimes, demonstrating that comparative conclusions in continual learning may depend strongly on the chosen fine-tuning setup.

\item \textbf{An optimization level interpretation of regime sensitivity.} Beyond comparative ranking changes, we analyze how regime choice affects the underlying optimization behavior. We show that deeper adaptation is associated with larger update magnitudes, greater forgetting, and a stronger relationship between the two. This provides an optimization level interpretation of regime sensitivity and helps explain why more permissive adaptation settings can produce different method behavior.

\item \textbf{Data and Code Availability.} Code, task order samples, and configuration files for all method-depth-dataset combinations used in this study are publicly available on our \href{https://github.com/PaulTiberiu/Trainable_Depth_CL}{\textcolor{blue}{GitHub}}. Beyond supporting reproducibility, this release is intended to serve as a benchmark for the regime-sensitivity perspective introduced in this work, allowing other researchers to test whether the ranking reversals we observe extend to additional methods, architectures, and datasets.

\end{itemize}

The study is structured as follows. We first review the related work in Section~\ref{sec: rel_work}. Next, Section~\ref{sec:our_approach} presents our theoretical framework, introducing the notion of adaptation regimes and analyzing how the trainable parameter subspace influences the optimization dynamics of continual learning. Section~\ref{sec:experiments} describes the experimental setup (Section~\ref{subsec:experimental_setup}) and the empirical evaluation, showing how fine-tuning regimes affect method rankings (Section~\ref{subsec:ranking_sensitivity}) and investigating the relationship between projected gradient magnitude and forgetting (Section~\ref{sec:grad_forgetting}). We then discuss the limitations of our study in Section~\ref{sec:limitations} before concluding, in Section~\ref{sec:conclusion}, with the main findings and their implications for continual learning evaluation. Finally, Section~\ref{sec: future_work} outlines future research directions motivated by the limitations identified in our study.

\section{Related Work}
\label{sec: rel_work}

\paragraph{Relation to Task Ordering and Benchmark Variability.}
Several works have observed that the order in which tasks are presented can substantially affect CL performance, yet this dependence is typically treated as noise rather than as a structured property of the problem \cite{wołczyk2021continualworldroboticbenchmark, Evron2023ContinualLID}. Benchmarks such as CLiMB \cite{Srinivasan2022CLiMBACB} highlight the variability induced by different task sequences, particularly in complex multimodal settings, but stop short of analyzing how this variability interacts with the training setup. Our work builds on this line of research and goes further by demonstrating that sensitivity to task ordering is itself regime-dependent: the same task ordering can yield different relative rankings of methods depending on which parameters are kept trainable, and this interaction must be treated as a joint design variable in CL evaluation. This aligns with findings that reordering tasks alone is sufficient to significantly alter forgetting outcomes even when the task set is held fixed \cite{Bell2022EffectOT}, and with related observations that order-induced variance persists across sequential benchmarks \cite{Kemker2018MeasuringCF}.

\paragraph{Relation to Parameter Isolation and Subspace Adaptation.}
A prominent line of work seeks to reduce inter-task interference by isolating or decomposing parameters across tasks \cite{Yoon2019ScalableAOA, Konishi2023ParameterLevelSFG, Liang2024InfLoRAILF}. These methods implicitly acknowledge that the trainable subspace matters for CL, treating subspace selection as a mechanism for preventing forgetting by ensuring that parameters important for past tasks are not modified when learning new ones. However, they do so as an algorithmic component rather than as an evaluation variable. In contrast, we fix the subspace throughout the CL sequence and study how its choice affects the comparative performance of standard methods across a range of trainable depths, from the last block of a ResNet-18 to full fine-tuning, treating it as an explicit axis of analysis rather than a design choice internal to any single algorithm. Related subspace-restriction strategies enforce orthogonality between successive task updates to curb interference \cite{Wang2023OrthogonalSL}, and recent geometric analyses show that the benefit of such restrictions is itself conditional on how task subspaces are positioned relative to one another \cite{Steele2026SubspaceGeometry}.

\paragraph{Relation to Plasticity, Stability, and the Role of Network Depth.}
The plasticity-stability trade-off has been studied from multiple angles, including the effect of network architecture and layer-wise learning dynamics on forgetting \cite{Raghu2019TransfusionUF, Achille2018CriticalLF}. It is well established that different layers of a deep network encode information at different levels of abstraction, with earlier layers capturing more general, transferable features and later layers encoding task-specific representations \cite{Yosinski2014HowTA}. This layer-wise asymmetry has direct implications for CL: the choice of which layers to update determines both how much plasticity the model retains for new tasks and how vulnerable previously acquired representations are to overwriting. While prior work has explored freezing early layers as a heuristic for reducing forgetting \cite{DeLange2022ACL}, no systematic study has examined how this choice interacts with and alters the comparative effectiveness of standard CL algorithms. Our work fills this gap by treating trainable depth as an explicit experimental variable rather than a fixed implementation detail. This connects to evidence that wider, shallower networks forget less catastrophically than their deeper counterparts \cite{Mirzadeh2022WideNN}, and to more recent architectural analyses showing that depth and width trade off plasticity against stability even under matched parameter budgets \cite{Lu2025RethinkingSP}.

\paragraph{Relation to Evaluation Methodology and Reproducibility in Continual Learning.}
Rigorous evaluation in CL has received increasing attention, with several works arguing that existing benchmarks and protocols suffer from inconsistencies that limit the reproducibility and comparability of results \cite{Farquhar2018TowardsRC, Vandeven2019ThreeSF, Biesialska2020ContinualLL}. Concerns include the choice of metrics, the influence of task ordering, and the sensitivity of conclusions to hyperparameter tuning, all of which can reverse method rankings across studies \cite{Wortsman2020SupermasksSC}. Our work contributes a complementary and underexplored dimension to this discussion: the fine-tuning regime itself. We show that even when task ordering, architecture, and hyperparameters are held fixed, varying only the trainable parameter subspace is sufficient to reverse the relative ranking of well-established CL methods. The broader framing that trainable subspace choice impacts the plasticity–stability trade-off and can alter comparative conclusions has been articulated in prior work: \cite{ren2024analyzingreducingcatastrophicforgetting} showed that different numbers of trainable parameters can lead to different CL method rankings in the context of LoRA versus full fine-tuning, while layer-depth freezing as a specific instantiation of subspace restriction has been explored in \cite{seo2025budgetedonlinecontinuallearning}. This finding suggests that the adaptation regime should be treated as a first-class evaluation variable in future CL benchmarking efforts, alongside the task sequence and the choice of algorithm. This echoes concerns that conventional hyperparameter-tuning protocols overestimate CL capability and fail to generalize across scenarios \cite{Cha2024HyperparametersCL}, as well as calls for standardized, fair-comparison frameworks across CL method families \cite{Masana2023ClassIncrementalSurvey}.

\paragraph{Differences from Prior Work.}
Unlike prior studies that examine task ordering, subspace selection, or network depth in isolation, we provide the first systematic study that jointly varies task ordering and trainable depth: from last-block to full fine-tuning, under fixed architecture and hyperparameters. This joint design reveals an interaction that isolated analyses cannot: the relative ranking of standard CL methods is not stable, but flips depending on the fine-tuning regime used. Where existing work treats task ordering and trainable subspace as separate, independently controllable factors, our results show they cannot be evaluated independently, establishing the adaptation regime as a first-class axis of CL benchmarking alongside task sequence and algorithm choice.

\section{Our Approach}
\label{sec:our_approach}

Our key claim is that an adaptation regime is not merely a choice of how many parameters remain trainable. It determines which update directions remain available throughout the task sequence and, therefore, changes the effective optimization signal seen by the learner. In continual learning, this matters because the learner must simultaneously fit the current task and preserve previously acquired knowledge.

\paragraph{Setup.}
We consider a sequence of tasks $\mathcal{T}_1, \dots, \mathcal{T}_T$ learned sequentially by a model with parameters $\theta \in \mathbb{R}^d$. At task $t$, the learner minimizes
\begin{equation}
\mathcal{J}_t(\theta)
=
\mathcal{L}_t(\theta)
+
\lambda \, \Omega_t(\theta; \theta_{1:t-1}),
\label{eq:cl_objective}
\end{equation}
where $\mathcal{L}_t$ is the current-task loss and $\Omega_t$ is the method-specific term used to preserve previously acquired knowledge. The purpose of Equation~\ref{eq:cl_objective} is simply to separate the two forces that coexist in continual learning: fitting the new task and preserving the past.

\paragraph{Adaptation Regime.}
An adaptation regime is defined by a fixed subset of trainable coordinates $S \subseteq \{1,\dots,d\}$, kept unchanged throughout the full continual learning run. We denote by $P_S$ the orthogonal projector that keeps the coordinates in $S$ and zeros out all others. In other words, once a regime is fixed, only the gradient components with respect to trainable parameters can induce updates.

Under a first-order method with step size $\eta>0$, the update at task $t$ becomes
\begin{equation}
\theta_t^{(m+1)}
=
\theta_t^{(m)}
-
\eta \, P_S \nabla \mathcal{J}_t(\theta_t^{(m)}),
\label{eq:projected_update}
\end{equation}
where $m$ indexes optimization steps within task $t$. Equation~\ref{eq:projected_update} is the key equation in this section: it makes explicit that the optimizer does not follow the full gradient of the objective, but only the component that lies in the trainable subspace.

\paragraph{Decomposing the Update Signal.}
To understand what changes across regimes, we decompose the projected gradient update in Equation~\ref{eq:projected_update} into the part coming from the current-task loss and the part coming from the preservation term:
\begin{equation}
g_t^{(S)}(\theta)
:=
P_S \nabla \mathcal{L}_t(\theta),
\qquad
r_t^{(S)}(\theta)
:=
P_S \nabla \Omega_t(\theta; \theta_{1:t-1}).
\label{eq:decomposition_terms}
\end{equation}
Using Equation~\ref{eq:cl_objective}, the projected gradient can then be written as
\begin{equation}
P_S \nabla \mathcal{J}_t(\theta)
=
g_t^{(S)}(\theta)
+
\lambda r_t^{(S)}(\theta).
\label{eq:signal_sum}
\end{equation}

Equations~\ref{eq:decomposition_terms} and~\ref{eq:signal_sum} isolate the two components of the projected update that every continual learning method must balance. The term $g_t^{(S)}$ is the update signal coming from the current task, while $r_t^{(S)}$ is the update signal induced by the preservation mechanism. A simple summary of their interaction is
\begin{equation}
\Gamma_t(S;\theta)
:=
\langle g_t^{(S)}(\theta), r_t^{(S)}(\theta)\rangle.
\label{eq:interaction}
\end{equation}
If $\Gamma_t(S;\theta)$ is positive, the two signals are locally compatible in the trainable subspace. If it is negative, fitting the current task and preserving previous knowledge compete for the same feasible directions.

\paragraph{Regime Dependence of the Update.}
The regime changes all three quantities above. It changes the current-task update signal because some gradient directions become inaccessible when parts of the network are frozen. It changes the preservation signal because the continual learning mechanism can only act through the same trainable coordinates. And it changes their interaction because alignment or conflict is evaluated only in feasible directions. This is the sense in which trainable depth changes the effective optimization signal, not just the number of trainable parameters.

\paragraph{A Minimal Optimization Consequence.}
Under standard smoothness assumptions, local progress depends on the norm of the projected update in Equation~\ref{eq:projected_update}, not on the norm of the full gradient. Concretely, if $\mathcal{J}_t$ is $L$-smooth and
\begin{equation}
\theta^+ = \theta - \eta P_S \nabla \mathcal{J}_t(\theta),
\label{eq:theta_plus}
\end{equation}
then for a sufficiently small step size,
\begin{equation}
\mathcal{J}_t(\theta^+)
\le
\mathcal{J}_t(\theta)
-
c_\eta \, \|P_S \nabla \mathcal{J}_t(\theta)\|^2
\qquad \text{for some } c_\eta > 0.
\label{eq:progress_statement}
\end{equation}

A proof is given in Appendix~\ref{appendix:proofs}. The role of Equation~\ref{eq:progress_statement} is limited but important: it shows that under a fixed regime, the quantity governing local progress is the projected update, not the full gradient. This is why our analysis focuses on projected quantities. Combining Equation~\ref{eq:signal_sum} with Equation~\ref{eq:progress_statement} shows that local progress depends on the current-task signal, the preservation signal, and their interaction inside the trainable subspace. Therefore, regime changes shift the update that actually drives learning.

\begin{figure*}[!t]
    \centering
    \begin{subfigure}{0.49\textwidth}
        \centering
        \includegraphics[width=\linewidth]{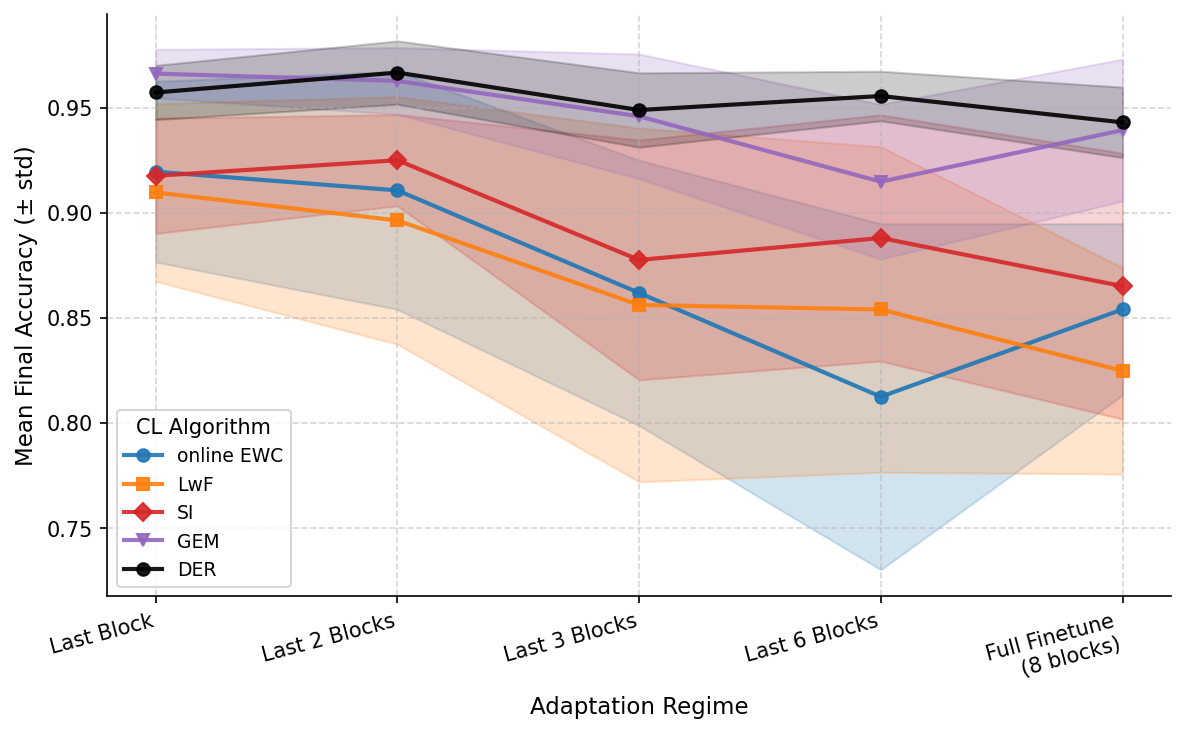}
        \caption{Final accuracy across regimes}
        \label{fig:fashion_line}        
    \end{subfigure}
    \hfill
    \begin{subfigure}{0.49\textwidth}
        \centering
        \includegraphics[width=\linewidth]{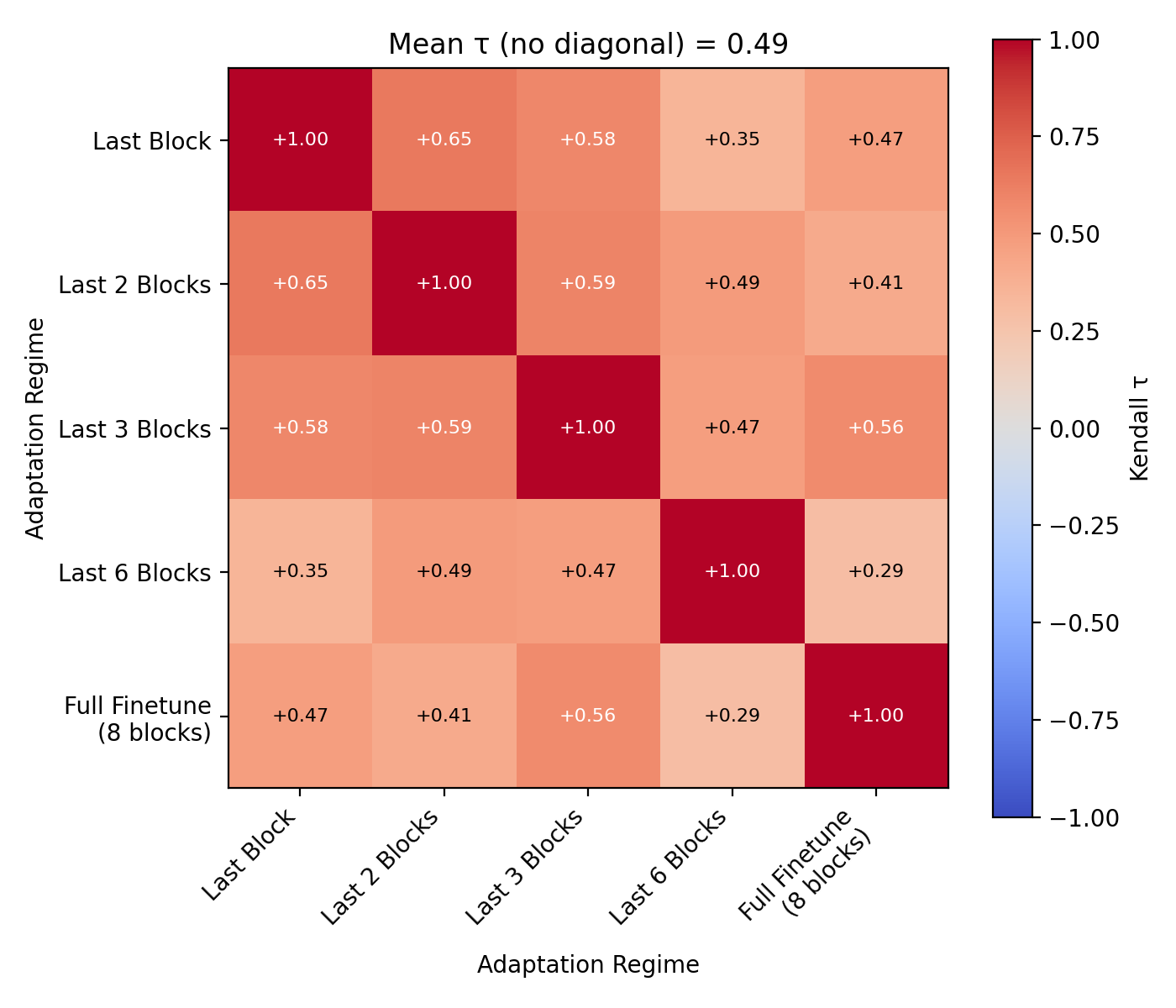}
        \caption{Mean Kendall's $\tau$ between regime specific rankings}
        \label{kendall_fashion_mnist}
    \end{subfigure}
    \begin{subfigure}{0.49\textwidth}
        \centering
        \includegraphics[width=\linewidth]{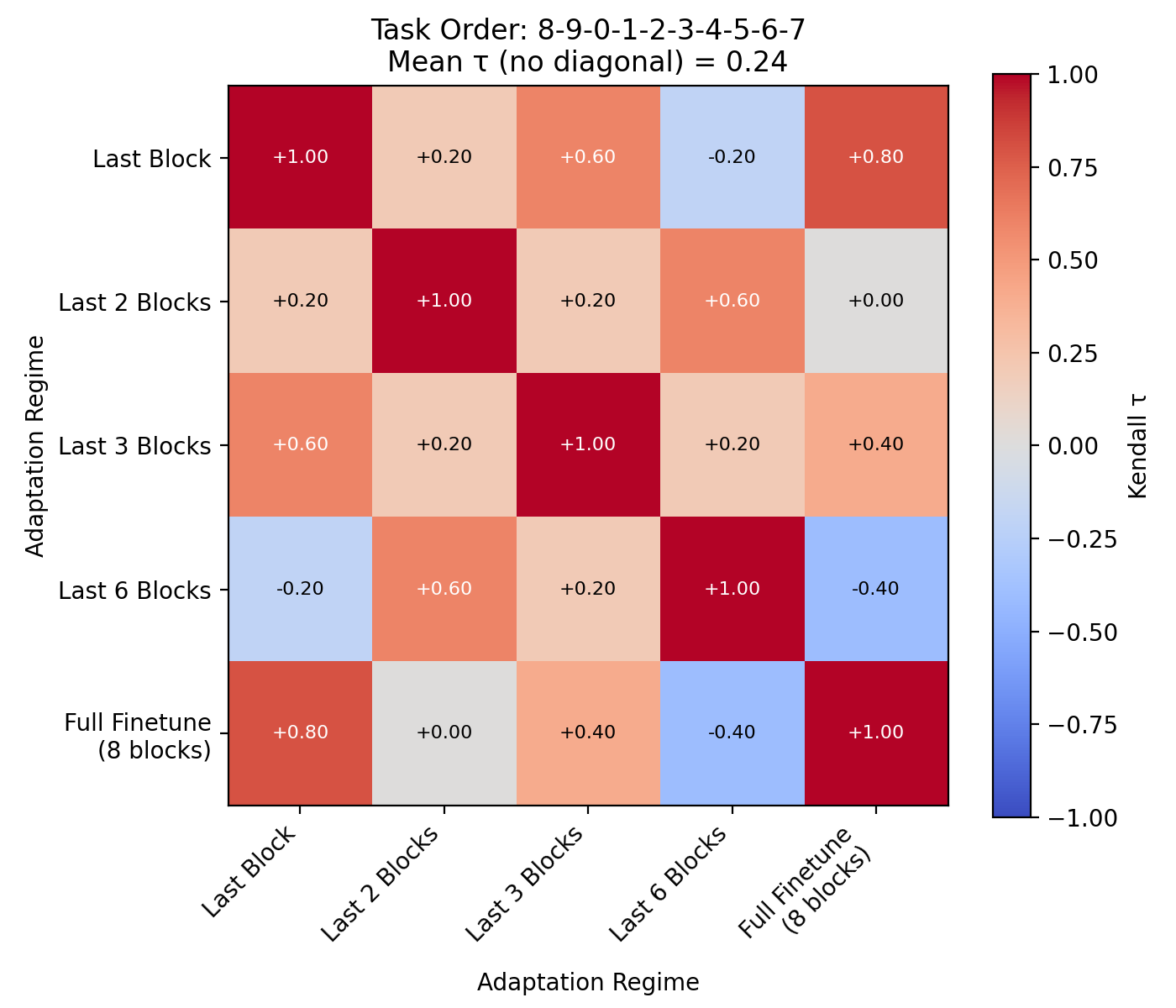}
        \caption{A task order with low ranking agreement across regimes}
        \label{fig:fashion_best}
    \end{subfigure}
    \hfill
    \begin{subfigure}{0.49\textwidth}
        \centering
        \includegraphics[width=\linewidth]{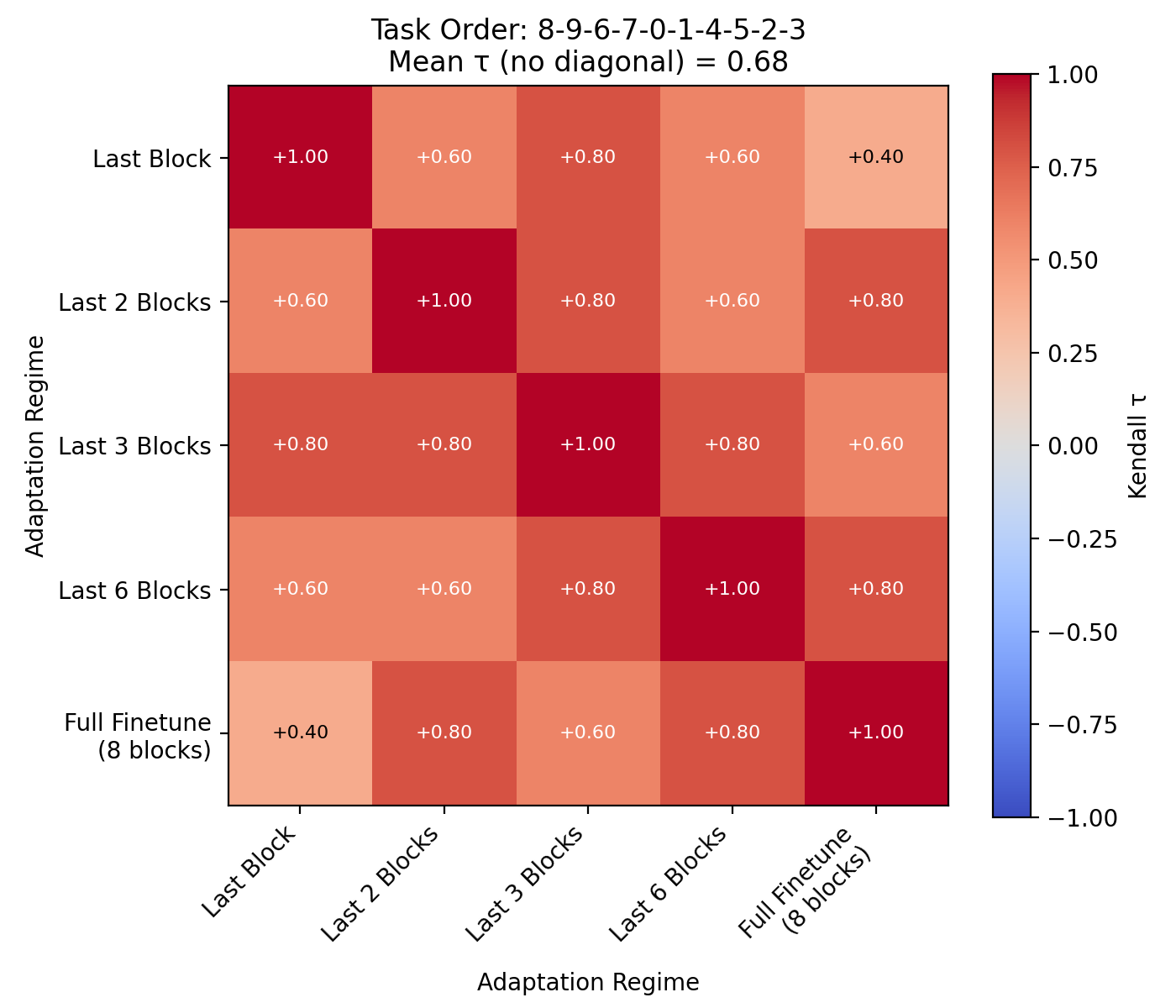}
        \caption{A task order with high ranking agreement across regimes}
        \label{fig:fashion_worst}
    \end{subfigure}
    \caption{Fine-tuning regime changes comparative conclusions (Fashion MNIST). (a) Standard continual learning methods exhibit different accuracy trends as the fine-tuning regime changes, with very large standard deviations. (b) Pairwise Kendall’s $\tau$ between rankings of CL algorithms under different fine-tuning regimes shows frequent changes in their relative ordering. (c,d) Two representative task orders illustrate that ranking stability can vary substantially, from low agreement across regimes to relatively high agreement. Together, these results show that trainable depth can alter the relative ranking of CL algorithms, even within the same benchmark. Reported values are mean and std for the final accuracy metrics, for the considered CL methods (online EWC, LwF, SI, GEM and DER) over $11$ sampled task orders.}
    \label{fig:main_results_fashion_mnist}
\end{figure*}

\begin{figure*}[t]
    \centering
    \begin{subfigure}{0.48\textwidth}
        \centering
        \includegraphics[width=\linewidth]{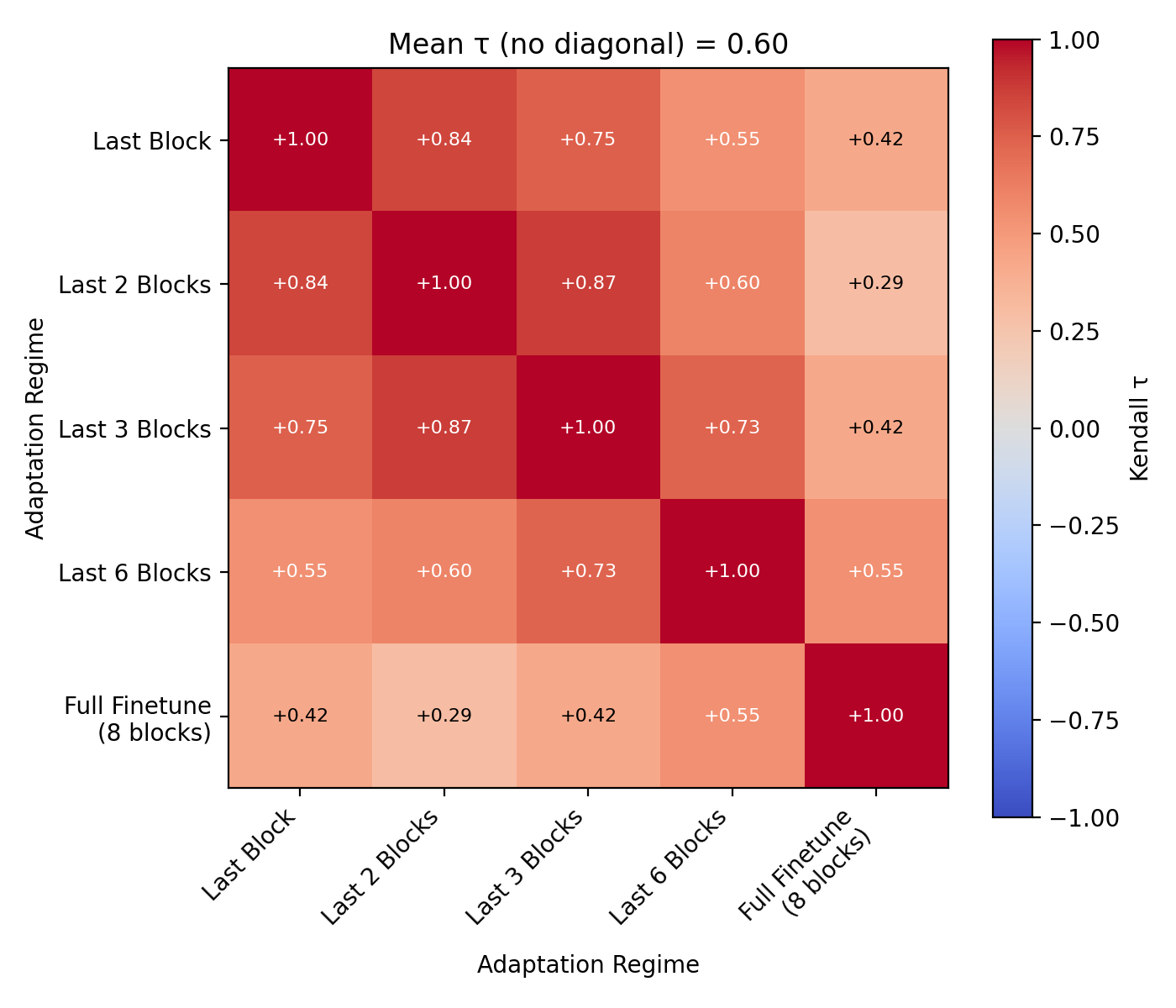}
        \caption{CIFAR-100}
        \label{kendall_cifar}
    \end{subfigure}
    \hfill
    \begin{subfigure}{0.48\textwidth}
        \centering
        \includegraphics[width=\linewidth]{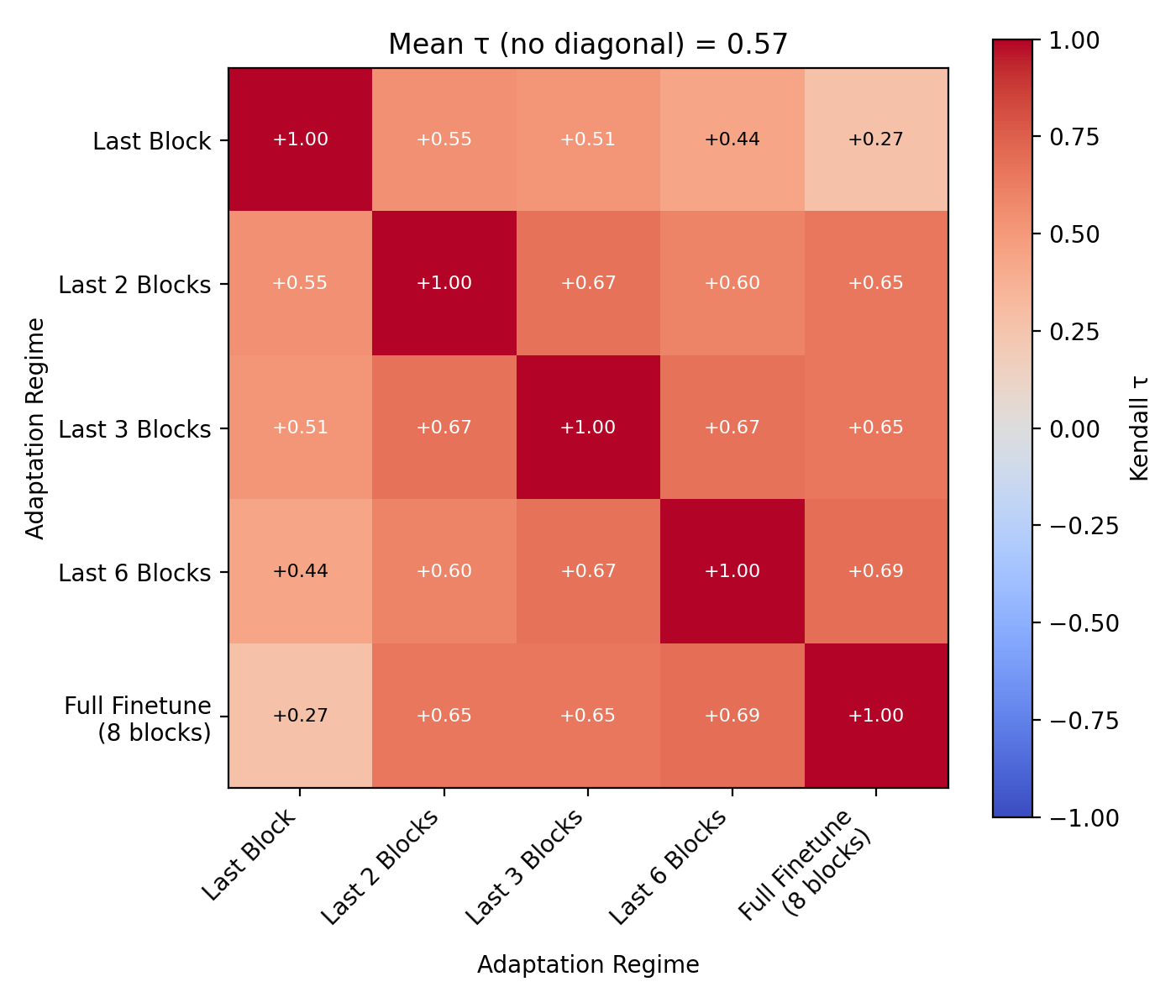}
        \caption{QMNIST}
        \label{kendall_qmnist}
    \end{subfigure}
    \hfill
    \begin{subfigure}{0.48\textwidth}
        \centering
        \includegraphics[width=\linewidth]{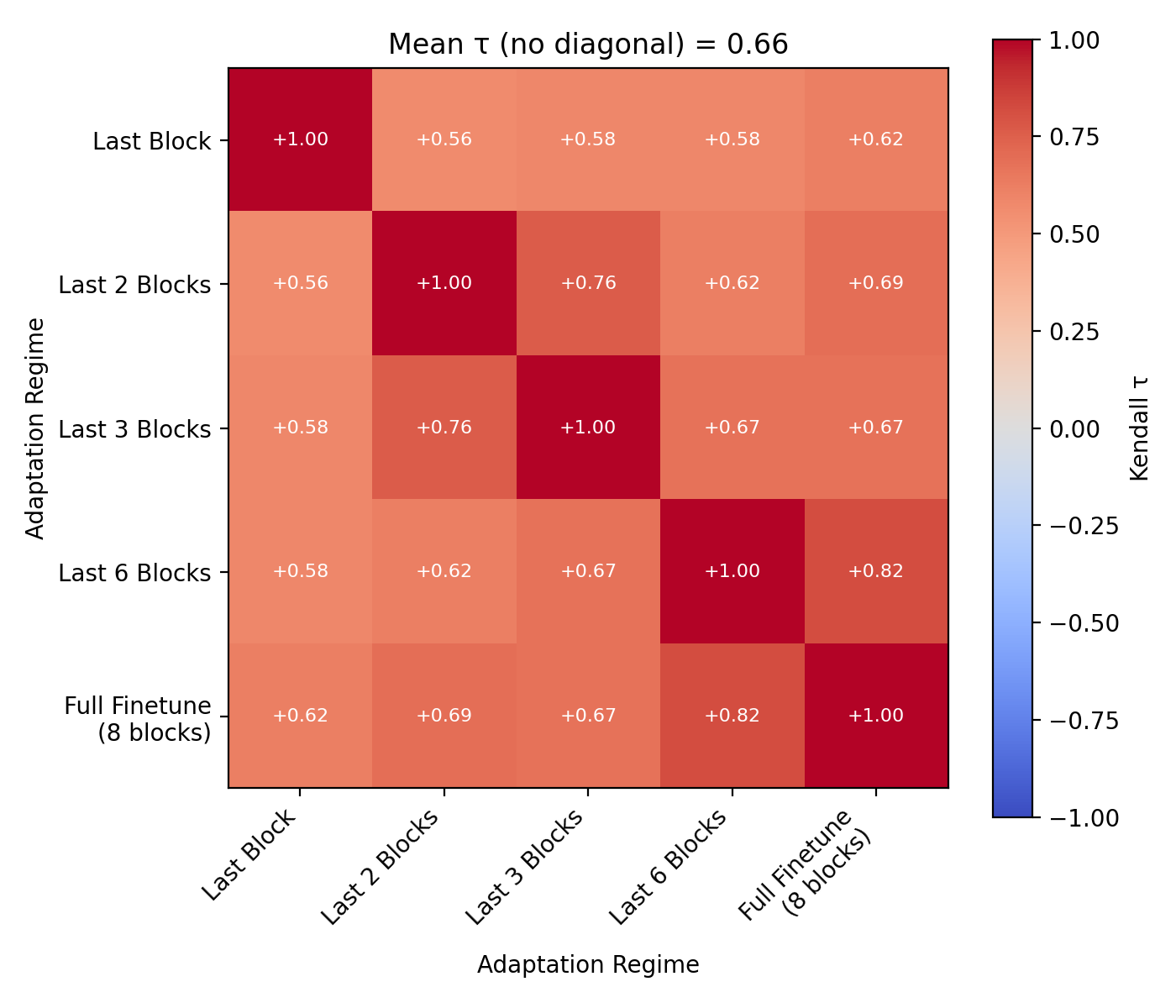}
        \caption{MNIST}
        \label{kendall_mnist}
    \end{subfigure}
    \hfill
    \begin{subfigure}{0.48\textwidth}
        \centering
        \includegraphics[width=\linewidth]{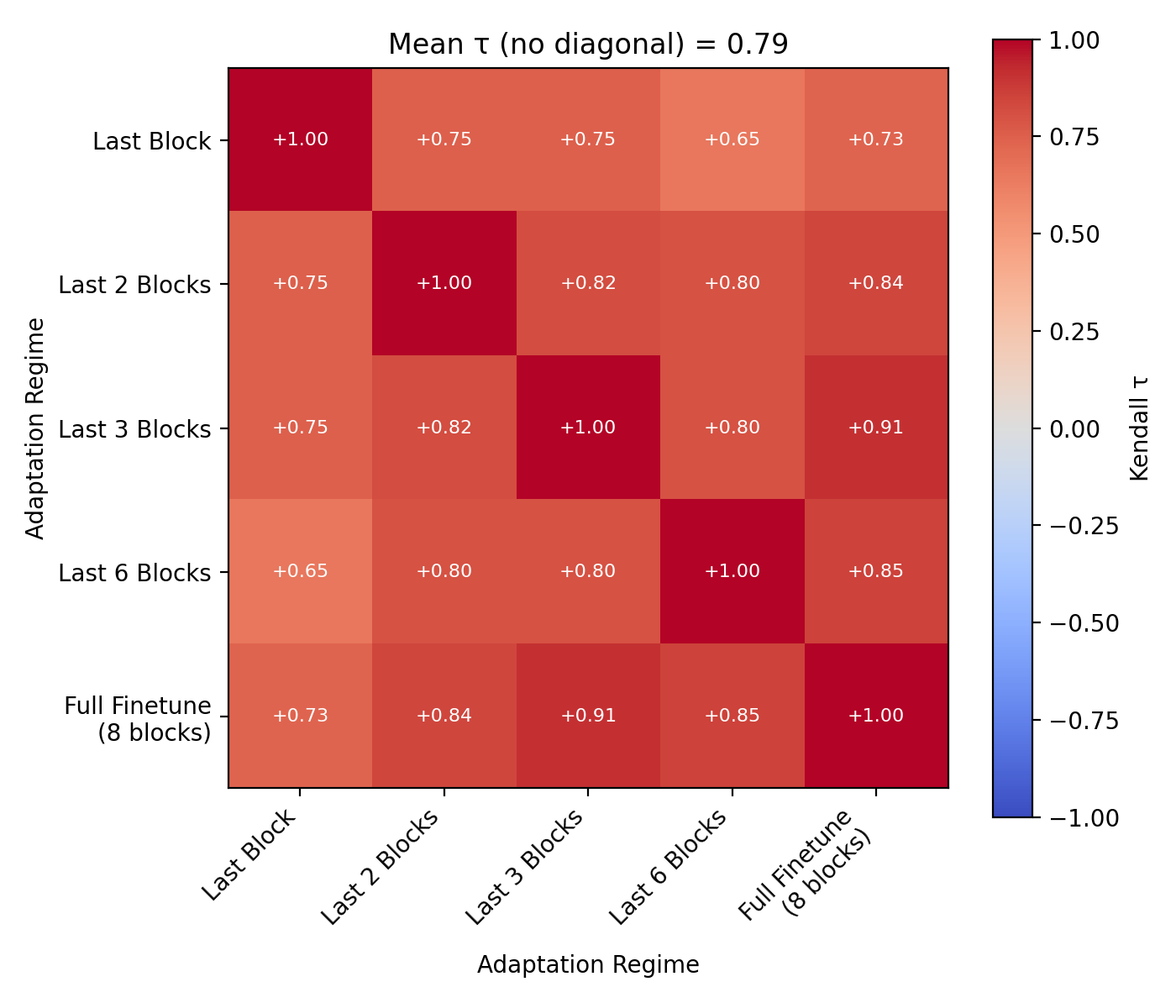}
        \caption{KMNIST}
        \label{kendall_kmnist}
    \end{subfigure}
    \caption{Ranking sensitivity persists across other benchmarks. Each panel shows the mean Kendall’s $\tau$ between regime-specific method rankings, averaged over $11$ randomly sampled task orders. Lower $\tau$ values indicate lower agreement between regimes in the ranking of the considered continual learning algorithms: online EWC, LwF, SI, GEM and DER. This disagreement is consistently observed across all considered benchmarks, namely Fashion MNIST, CIFAR-100, MNIST, QMNIST, and KMNIST, indicating that the relative ordering of these algorithms depends strongly on the fine-tuning regime used throughout the continual learning run.}
\label{fig:main_results_all_datasets}
\end{figure*}

\paragraph{From Theory to Empirical Validation.}
The analysis above yields direct empirical predictions, which we test in Section~\ref{sec:experiments}. At the comparative level, Figures~\ref{fig:main_results_fashion_mnist} and~\ref{fig:main_results_all_datasets} show that method rankings are not preserved across adaptation regimes, both within a single benchmark and across benchmark families. This is the observable consequence of changing the feasible update subspace through $P_S$: once the effective optimization signal changes, different CL methods need not retain the same relative ordering. To probe the mechanism behind this effect, we then analyze measurable projected quantities. In particular, for each task, order, method, and regime, we consider the magnitude of the projected current-task signal $\|g_t^{(S)}\|$, the magnitude of the projected preservation signal $\|r_t^{(S)}\|$, their interaction $\Gamma_t(S;\theta)$, and the norm of the resulting projected update, $\|P_S \nabla \mathcal{J}_t(\theta)\|^2$. Figure~\ref{fig:grad_forgetting} examines one concrete instance of this mechanism by relating gradient magnitude to forgetting, showing that both increase with trainable depth and that their coupling becomes stronger in more permissive regimes.

\paragraph{From Regime Sensitivity to Ranking Changes.}
Different continual learning methods induce different preservation signals $r_t^{(S)}$, and their relative effectiveness depends on how these signals interact with the current-task signal $g_t^{(S)}$ inside the trainable subspace. In deep networks, changing the regime alters not only the size of this subspace, but also where adaptation is allowed to occur. Shallow regimes restrict updates to late layers, favoring feature reuse and local adjustment, whereas deeper regimes expose more representation-changing directions and more opportunities for interference. Consequently, varying trainable depth shifts the effective balance between plasticity and stability, which can alter the method's behavior and comparative rankings across regimes.

\section{Experiments}
\label{sec:experiments}

We organize the experiments to validate that trainable depth is a consequential evaluation variable in continual learning and to characterize the mechanisms underlying this sensitivity. We first describe the common experimental setup, including the datasets, CL methods, fine-tuning regimes, and evaluation protocol (Sec.~\ref{subsec:experimental_setup}). We then test whether changing only the fine-tuning regime alters comparative conclusions (Sec.~\ref{subsec:ranking_sensitivity}). Next, we examine how gradient magnitude and forgetting covary across regimes to better understand what drives the observed instability of rankings  (Sec.~\ref{sec:grad_forgetting}).

\subsection{Experimental Setup}
\label{subsec:experimental_setup}

\paragraph{Continual Learning Setting.}
We study task incremental continual learning over a sequence of $T$ tasks, $\{\mathcal{T}_1, \ldots, \mathcal{T}_T\}$, where task identity is available at both training and test time. Each task contains $C$ disjoint classes. We use a shared
backbone together with a global $T \times C$ classifier, and at inference time we select the logits corresponding to the current task. Task boundaries are assumed to be known throughout training. No data from previous tasks are retained unless explicitly required by the method, as in GEM.

\vspace{-2.5mm}

\paragraph{Model and Fine-tuning Regimes.}
Our backbone is a randomly initialized ResNet-18~\cite{he2015deepresiduallearningimage}. We use ResNet-18 as the backbone because its fixed, well-established architecture provides a controlled and reproducible testbed for studying the sensitivity of continual learning methods to the amount of trainable depth. In particular, its modular residual-block structure allows the trainable subspace to be varied unambiguously while keeping the overall architecture unchanged, thereby isolating trainable depth as the primary experimental variable. To study the impact of trainable depth, we define 5 fine-tuning regimes by varying the number of trainable residual blocks: training only the last block, the last 2 blocks, the last 3 blocks, the last 6 blocks, or all 8 blocks (i.e., full fine-tuning). These regimes were selected to span a broad spectrum of adaptation flexibility, from highly restricted adaptation of only the deepest layers to full-network training, while covering different portions of the ResNet-18 hierarchy and therefore different levels of feature abstraction. This design allows us to examine whether the relative performance of continual learning methods is preserved as the amount of plasticity in the backbone changes. All remaining parameters are frozen throughout the CL sequence. This design lets us span a spectrum from highly constrained to fully flexible adaptation, with the entire network allowed to drift over time.

\paragraph{Datasets.}
We evaluate on 5 benchmarks spanning 2 dataset families: 1 natural image benchmark, CIFAR-100 and 4 MNIST-based benchmarks, namely MNIST, Fashion MNIST, KMNIST, and QMNIST. Each dataset is partitioned into $T$ disjoint tasks with $C$ classes per task, following standard task incremental protocols. For the MNIST-based benchmarks, we use $T=5$ tasks with $C=2$ classes per task, while CIFAR-100 is divided into $T=10$ tasks with $C=10$ classes per task. The MNIST-based datasets enable controlled comparisons across benchmarks with similar structure but different visual characteristics, while CIFAR-100 provides a more challenging setting with richer visual features and greater inter-task diversity.

\paragraph{Baselines.}
We compare 5 standard continual learning methods: online EWC~\cite{schwarz2018progresscompressscalable}, LwF~\cite{li2017learningforgetting}, SI~\cite{zenke2017continuallearningsynapticintelligence}, GEM~\cite{lopezpaz2022gradientepisodicmemorycontinual}, and DER~\cite{yan2021derdynamicallyexpandablerepresentation}. These methods cover several major CL families, including those based on regularization, distillation, gradient projection, and dynamic architecture methods. The comparison focuses exclusively on continual learning methods, and therefore a naive fine-tuning baseline was not considered necessary for the primary analysis. Such a baseline would correspond to training on each task without a continual learning mechanism and would primarily quantify catastrophic forgetting, whereas our objective is to assess whether the relative ranking of established continual learning strategies remains stable across different trainable-depth regimes. The baseline comparison is thus performed directly among CL methods under identical task orders, datasets, and backbone configurations.

\paragraph{Evaluation Protocol.}
For each dataset, fine-tuning regime, and CL method, we evaluate $N_{\text{ord}} = 11$ task orders: 1 canonical ordering and 10 randomly sampled orderings. The same set of task orders is reused across all methods and regimes to ensure a fair comparison. Our main goal is not only to compare raw performance, but also to assess whether the relative ranking of methods is preserved as trainable depth changes. To this end, we compute Kendall’s $\tau$ rank correlation between the method rankings induced by different fine-tuning regimes. High $\tau$ indicates that two regimes produce similar comparative conclusions, whereas low $\tau$ indicates that changing the fine-tuning regime alters the relative ordering of methods.

\vspace{-1.5mm}

\paragraph{Evaluation Metrics.}
We report two standard CL metrics. The first is average accuracy, defined as the mean final accuracy over all tasks,
$
\bar{A} = \frac{1}{T}\sum_{i=1}^{T} A_T^{(i)},
$
where $A_T^{(i)}$ denotes the accuracy on task $i$ after training on the full sequence of $T$ tasks. The second is average forgetting, which measures the extent to which performance on previous tasks deteriorates over time,
$
\bar{F} = \frac{1}{T-1}\sum_{i=1}^{T-1} \left( \max_{t \leq T} A_t^{(i)} - A_T^{(i)} \right).
$
Higher forgetting values indicate a larger loss of previously acquired knowledge by the end of training, whereas values closer to zero indicate better retention.

\begin{figure*}[t]
\centering
    \begin{subfigure}{0.48\textwidth}
        \centering
        \includegraphics[width=\linewidth]{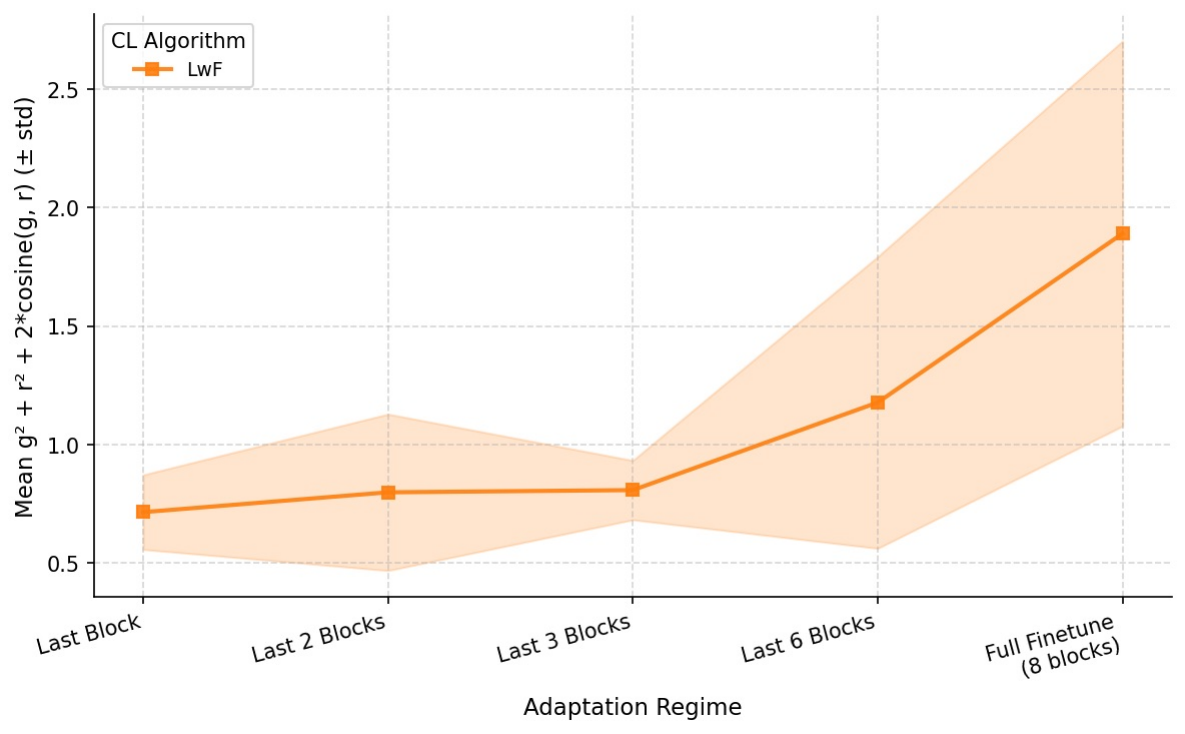}
        \caption{Gradient magnitude increases with trainable depth}
    \end{subfigure}
    \hfill
    \begin{subfigure}{0.48\textwidth}
        \centering
        \includegraphics[width=\linewidth]{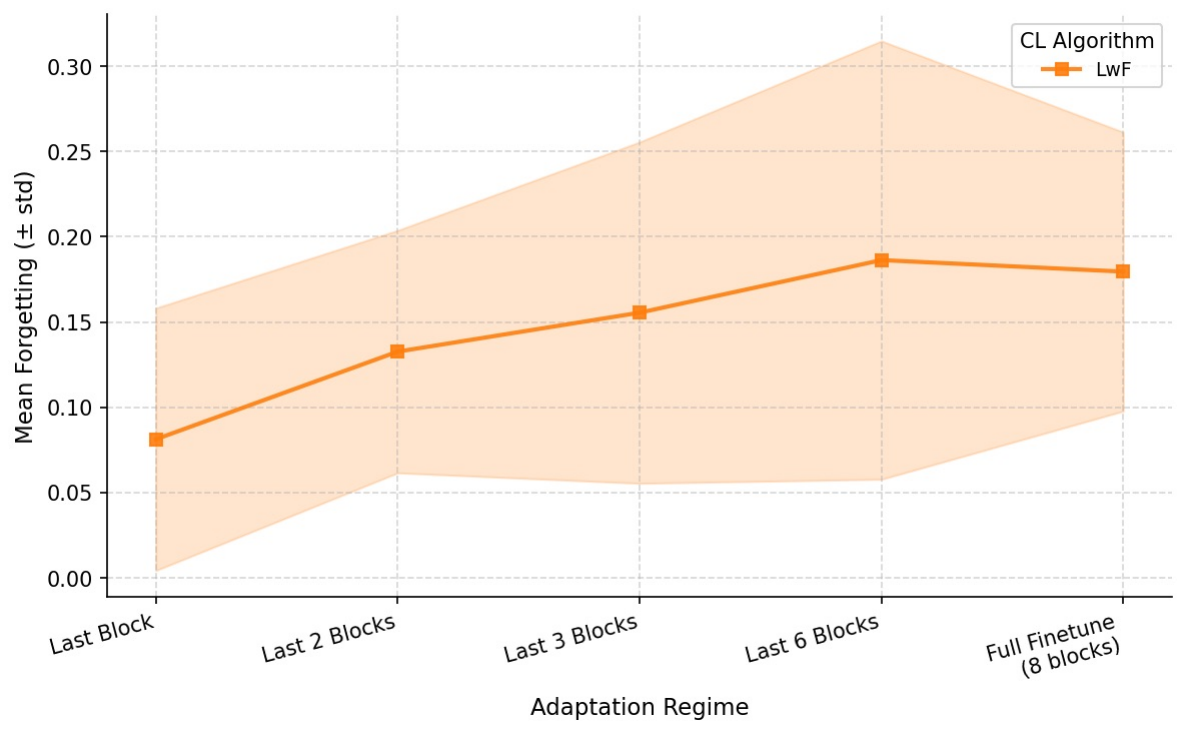}
        \caption{Forgetting increases with trainable depth}
    \end{subfigure}
    \begin{subfigure}{0.48\textwidth}
        \centering
        \includegraphics[width=\linewidth]{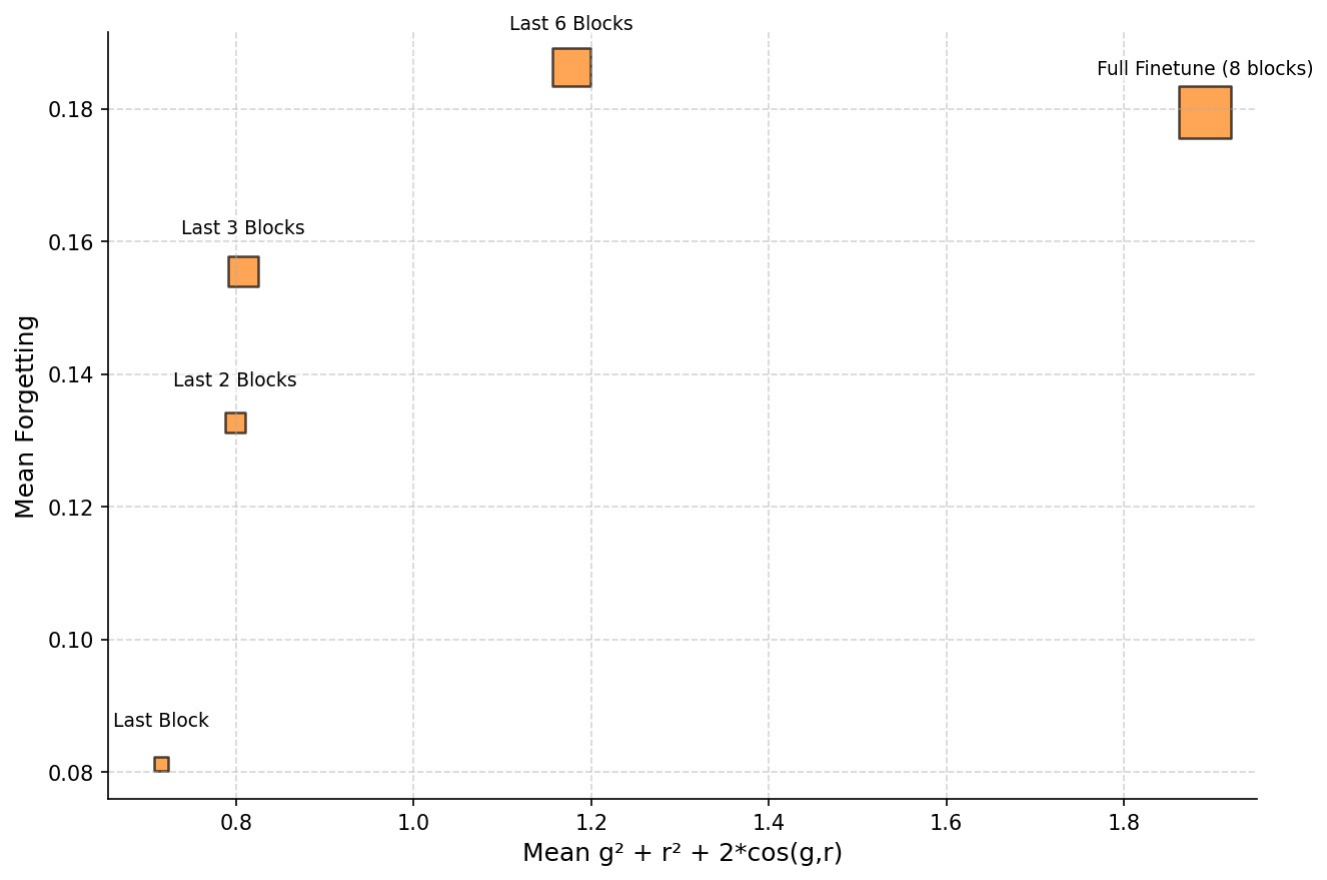}
        \caption{Across fine-tuning regimes, larger gradient magnitudes are associated with higher forgetting}
    \end{subfigure}
       \hfill
    \begin{subfigure}{0.48\textwidth}
        \centering
        \includegraphics[width=\linewidth]{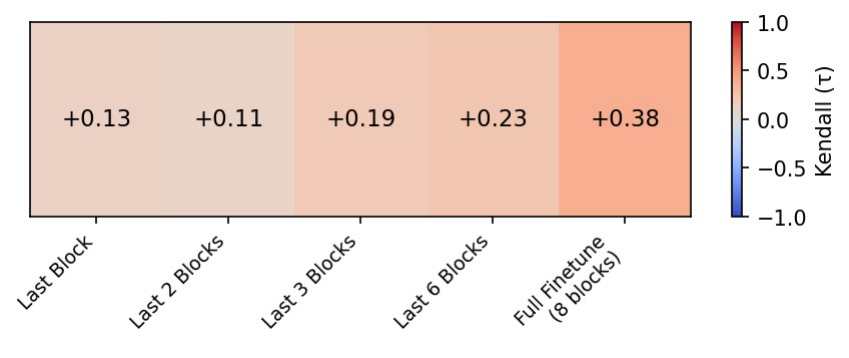}
        \caption{The Kendall $\tau$ correlation between gradient magnitude and forgetting becomes stronger in more permissive regimes}
    \end{subfigure}
    \caption{Gradient magnitude and forgetting both increase with trainable depth, but the main effect goes beyond these parallel trends. As adaptation becomes more permissive, forgetting becomes increasingly tied to gradient magnitude. (a) Mean gradient magnitude increases as more parameters are made trainable. (b) Average forgetting also increases with trainable depth. (c) Across regimes, larger gradient magnitudes are associated with higher forgetting. (d) The Kendall $\tau$ correlation between gradient magnitude and forgetting becomes stronger in more permissive regimes, indicating that gradient magnitude becomes increasingly informative of forgetting.}
    \label{fig:grad_forgetting}
\end{figure*}

\paragraph{Analysis of Ranking Stability.}
Beyond raw performance, our central diagnostic is the stability of method rankings across fine-tuning regimes. For each task order, we rank CL methods according to their average accuracy within each regime, and then compute Kendall’s $\tau$~\cite{kendall1938} for every pair of regimes. This provides a direct measure of how consistently methods are ordered as trainable depth varies. A value of $\tau = 1$ indicates perfect agreement, meaning that two regimes induce the same method ranking, while $\tau = -1$ indicates a complete reversal. Values near zero indicate weak agreement, suggesting that conclusions drawn under 1 regime may not transfer reliably to another. To summarize this effect at the dataset level, we average the pairwise $\tau$ values across all 11 task orders, yielding a single matrix that captures rank stability over the entire evaluation. In the gradient-based analysis of Figure~\ref{fig:grad_forgetting}, we use the same statistic to quantify how consistently regimes are ordered when ranked by gradient magnitude and by forgetting.

\paragraph{Hyperparameters}

\begin{table}[!h]
\caption{\textbf{General training hyperparameters used across all experiments.}}
\label{tab:training_hyperparameters}
\footnotesize
\begin{tabularx}{\columnwidth}{@{}lX@{}}
\toprule
\textbf{Hyperparameter} & \textbf{Value} \\
\midrule
Optimizer & Adam \\
\midrule
Initial learning rate & $10^{-3}$ \\
Minimum learning rate (Cosine Annealing) & $10^{-5}$ \\
\midrule
Batch size & $256$ \\
\midrule
Epochs per task (CIFAR-100) & $20$ \\
Epochs per task (MNIST variants) & $3$ \\
\midrule
Early stopping patience & $5$ epochs \\
Minimum improvement ($\Delta$) & $10^{-4}$ \\
\bottomrule
\end{tabularx}
\end{table}

\begin{table}[!h]
\caption{\textbf{Algorithm-specific hyperparameters for the continual learning methods.}}
\label{tab:algorithm_hyperparameters}
\footnotesize
\begin{tabularx}{\columnwidth}{@{}llX@{}}
\toprule
\textbf{Method} & \textbf{Hyperparameter} & \textbf{Value} \\
\midrule

DER & $\alpha$ & $1.0$ \\
\midrule

\multirow{2}{*}{SI}
& $c$ & $100.0$ \\
& $\xi$ & $0.1$ \\
\midrule

\multirow{4}{*}{Online EWC}
& $\lambda$ & $5000.0$ \\
& $\gamma$ & $1.0$ \\
& Fisher samples & $200$ \\
& Fisher batch size & $1$ \\
\midrule

\multirow{3}{*}{LwF}
& $\lambda$ & $1.0$ \\
& Temperature $T$ & $2.0$ \\
& Warm-up epochs & $10$ (CIFAR), $3$ (MNIST) \\
\midrule

\multirow{2}{*}{GEM}
& Memory (samples/task) & $200$ \\
& QP $\epsilon$ & $10^{-5}$ \\
\midrule

\multirow{2}{*}{ER}
& Buffer size & $2000$ \\
& Replay batch size & $256$ \\
\bottomrule
\end{tabularx}

\end{table}

To facilitate reproducibility, we report all hyperparameters used throughout our experiments. Table~\ref{tab:training_hyperparameters} summarizes the general training configuration, while Table~\ref{tab:algorithm_hyperparameters} lists the algorithm-specific hyperparameters. Unless otherwise stated, these hyperparameters are kept fixed across datasets, task orders, and fine-tuning regimes, ensuring that observed differences are attributable to the adaptation regime rather than changes in optimization or method configuration.

\subsection{Trainable Depth Reorders Continual Learning Methods}

\label{subsec:ranking_sensitivity}

We investigate whether the ranking of standard CL methods is stable across fine-tuning regimes or whether it depends on the trainable depth used throughout the CL run. This question is important because comparative CL evaluations are typically performed under a single adaptation setting, implicitly assuming that the resulting ranking is method-specific rather than configuration-dependent. To test this assumption, we evaluate 5 standard methods (i.e., online EWC, LwF, SI, GEM and DER) under multiple fine-tuning regimes and compare the resulting method rankings using pairwise Kendall’s $\tau$.

Figure~\ref{fig:main_results_fashion_mnist} first illustrates the effect on Fashion MNIST. Varying trainable depth changes both the methods' accuracy profiles and their induced rankings, with representative task orders showing that the same set of methods can exhibit either relatively stable or strongly unstable rankings depending on the sequence. Figure~\ref{fig:main_results_all_datasets} then shows that this phenomenon persists across all considered benchmarks when Kendall’s $\tau$ is averaged over $11$ randomly sampled task orders per dataset.

Across datasets, regime-specific rankings exhibit only partial agreement, with mean $\tau$ ranging from $0.49$ on Fashion MNIST to $0.79$ on KMNIST. The strongest disagreements consistently occur between the most restrictive and most permissive regimes, confirming that trainable depth is a consequential evaluation variable. At the dataset level, two patterns emerge.
Ranking instability remains evident, with method crossovers appearing across task orders even when mean accuracy curves indicate only minor performance differences.

Taken together, these results show that comparative conclusions in CL are sensitive to the fine-tuning regime under which they are drawn. Therefore, trainable depth should be treated as a first-class evaluation variable when comparing CL methods.

\begin{figure*}[t]
\centering
    \begin{subfigure}{0.48\textwidth}
        \centering
        \includegraphics[width=\linewidth]{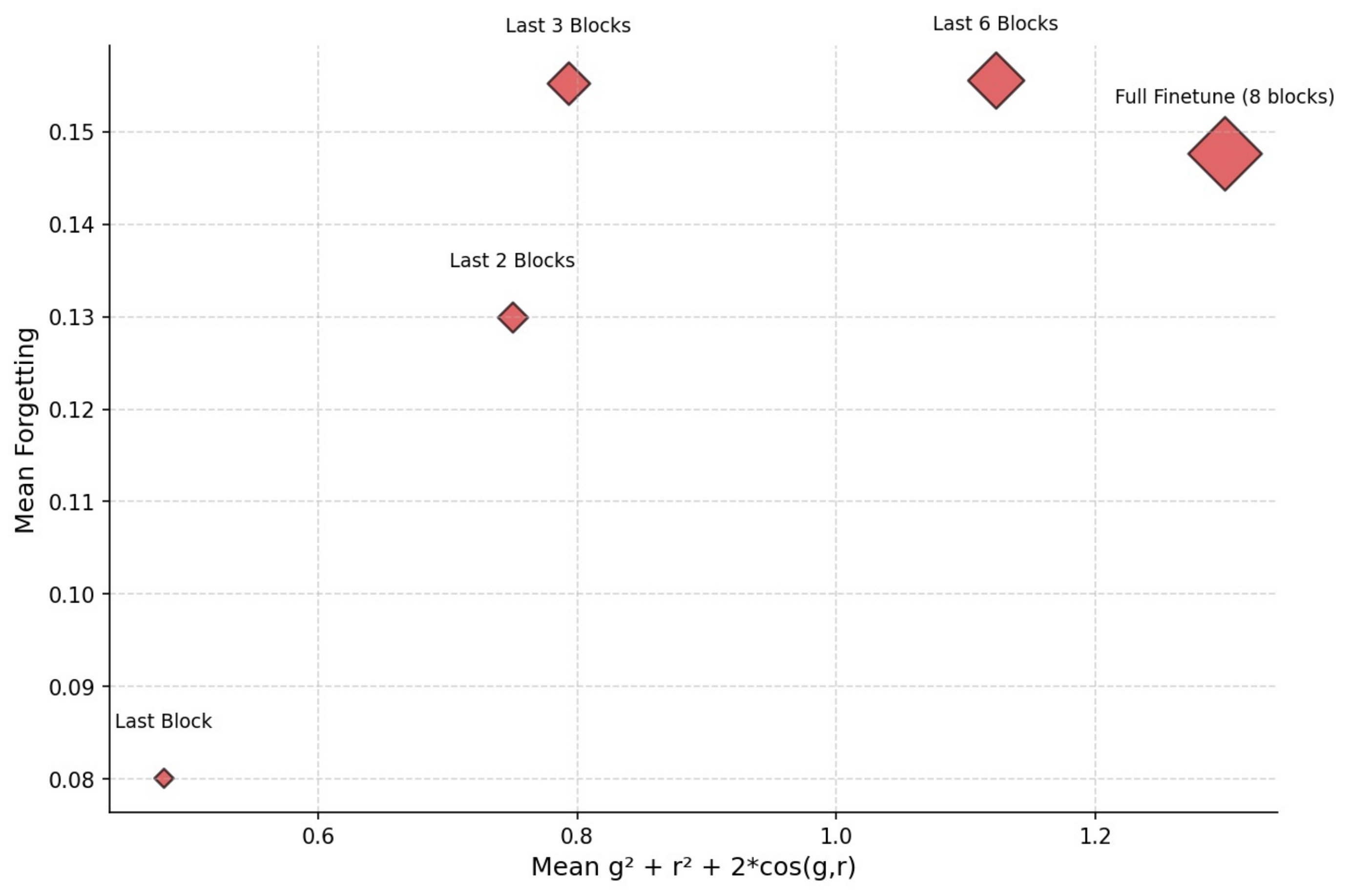}
        \caption{SI}
    \end{subfigure}
    \hfill
    \begin{subfigure}{0.48\textwidth}
        \centering
        \includegraphics[width=\linewidth]{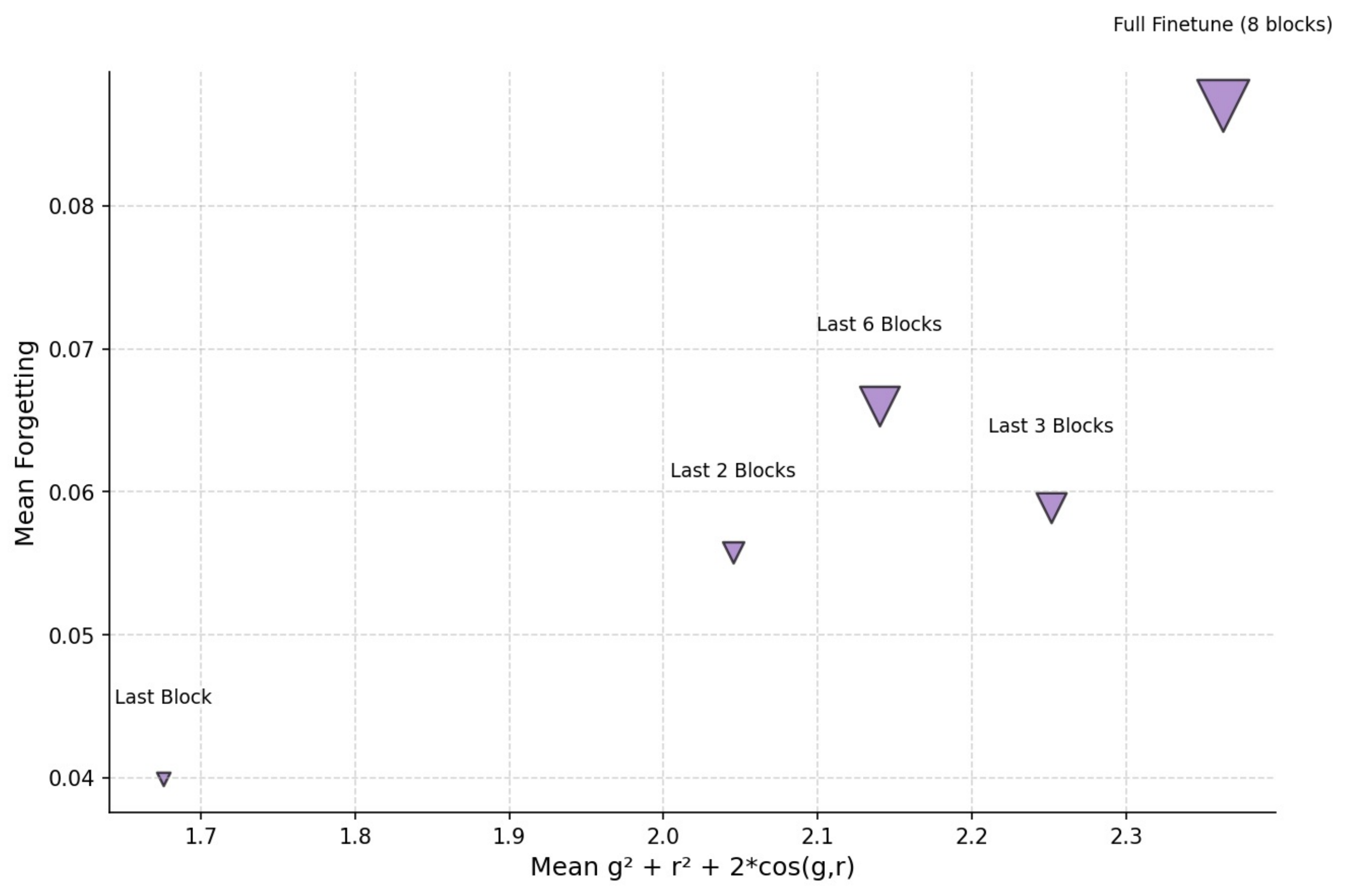}
        \caption{GEM}
    \end{subfigure}
    \begin{subfigure}{0.48\textwidth}
        \centering
        \includegraphics[width=\linewidth]{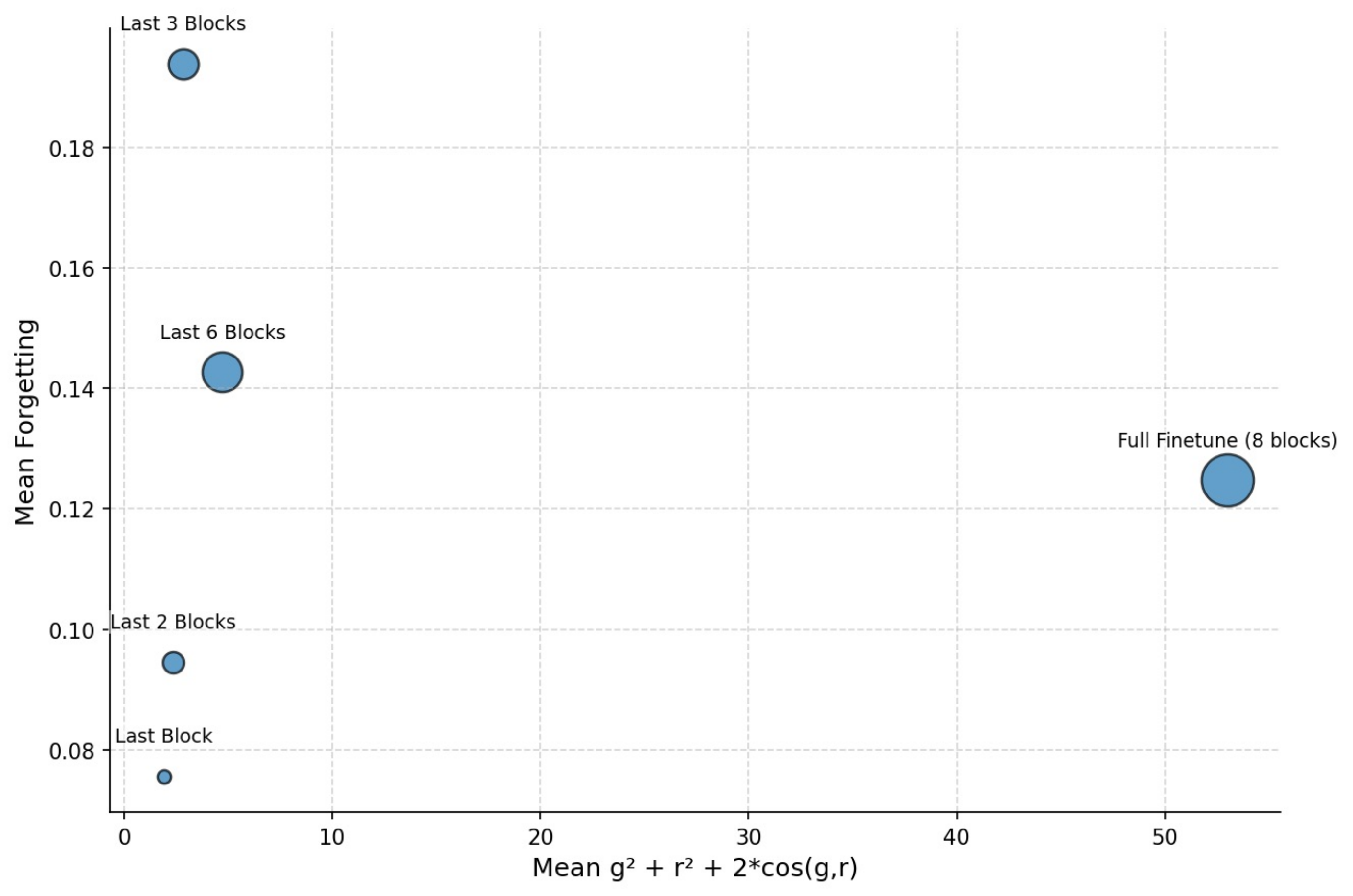}
        \caption{online EWC}
    \end{subfigure}
       \hfill
    \begin{subfigure}{0.48\textwidth}
        \centering
        \includegraphics[width=\linewidth]{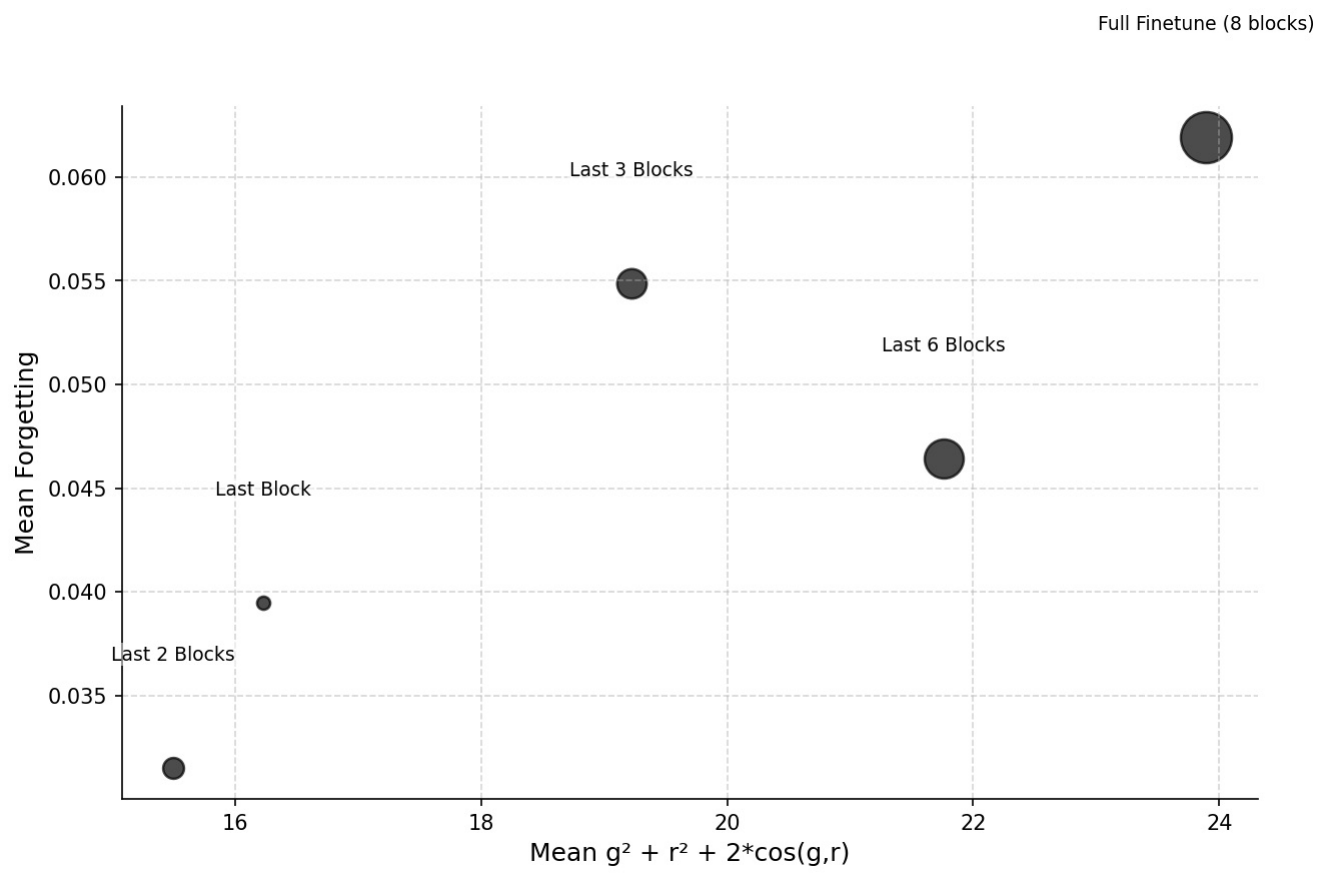}
        \caption{DER}
    \end{subfigure}
    \caption{Extension to Figure \ref{fig:grad_forgetting}c. (a) For SI, the relationship is clean and monotonic, following the exact same pattern as LwF. (b) online EWC follows a similar overall direction but less monotonically, with intermediate regimes exhibiting higher forgetting than expected from a strictly increasing trajectory. (c) GEM presents a different picture: forgetting still increases with depth, but the gradient norm values span a remarkably narrow range compared to online EWC, as expected from GEM’s gradient projection mechanism, which structurally suppresses update magnitudes across all configurations. One local inversion occurs, where an intermediate regime shows a smaller norm but higher forgetting than the next deeper regime. (d) DER follows a broadly increasing trend between gradient norm and forgetting, but exhibits one local inversion: Last 3 Blocks shows higher forgetting than Last 6 Blocks despite a lower norm, likely because DER's logit-replay constraint conflicts more sharply with gradient updates when fewer blocks are unfrozen. The norm range across configurations is moderate: wider than GEM but narrower than online EWC, consistent with DER's soft output-matching regularization. Full fine-tuning anchors the upper-right extreme as expected.}
    \label{fig:grad_forgetting_2}
\end{figure*}

\subsection{Gradient Magnitude and Forgetting Couple More Tightly in Permissive Regimes}
\label{sec:grad_forgetting}

Our theoretical analysis suggests that the effect of an adaptation regime is not exhausted by the number of trainable parameters. By changing the projector $P_S$, the regime changes the set of feasible update directions and, therefore, the effective optimization signal that can propagate through the network. In more permissive regimes, the learner has access to a larger trainable subspace, which creates more room for representational change. In continual learning, this additional freedom should have two observable consequences: stronger update magnitudes and a higher risk that these updates interfere with previously acquired knowledge.

To test this prediction, we analyze the relationship between gradient magnitude and forgetting across fine-tuning depths using a single CL method, namely LwF, on a single dataset, namely Fashion MNIST. We focus on LwF because its training objective is most directly aligned with the decomposition in Equation~\ref{eq:signal_sum}, where the projected gradient of the full objective is written as the sum of a current task term and a preservation term. In LwF, the preservation mechanism enters as an explicit distillation loss added to the current task loss, so the resulting update is naturally described by the projected objective gradient $\|P_S \nabla J_t(\theta)\|^2$. This makes LwF a particularly clean setting for studying how the magnitude of updates relates to final forgetting across adaptation regimes.

For each regime, we compute a summary measure of the projected gradient magnitude during training and relate it to the final average forgetting at the end of the task sequence. We then examine how these two quantities vary across regimes and how consistently regimes are ordered when ranked by gradient magnitude and by forgetting.

Figure~\ref{fig:grad_forgetting} supports this view. As trainable depth increases, mean gradient magnitude rises (Figure~\ref{fig:grad_forgetting}a) and average forgetting rises as well (Figure~\ref{fig:grad_forgetting}b). However, the main result goes beyond these parallel trends. Across regimes, larger gradient magnitudes are associated with higher forgetting (Figure~\ref{fig:grad_forgetting}c), and this association becomes progressively stronger in more permissive regimes, as shown by the increasing Kendall $\tau$ correlation in Figure~\ref{fig:grad_forgetting}d. For results and comments on the remaining evaluated methods, we refer to Figure~\ref{fig:grad_forgetting_2}, where the relationship between $\|g\|^2 + \|r\|^2 + 2\cos(g, r)$ and mean forgetting holds consistently across the majority of evaluated methods and adaptation regimes. We also evaluated the metric $|g|^2 + |r|^2 + 2|g||r|\cos(g, r)$, which showed a very similar relationship with forgetting, although the observed correlations were generally not as strong.

This pattern is consistent with the interpretation developed in Section~\ref{sec:our_approach}. When the regime is restrictive, the optimizer is confined to a limited set of feasible update directions, and the resulting dynamics are strongly shaped by the prior imposed by the frozen backbone. Under shallow adaptation, this structural prior limits both update flow and the scope of interference. As the regime becomes more permissive, optimization is less constrained by the prior and increasingly shaped by task-driven gradients, making the method's behavior more regime-dependent. In this sense, deeper adaptation does not merely increase the extent to which the network changes; it also makes those changes more consequential for previously learned tasks.

\section{Limitations}
\label{sec:limitations}

\paragraph{Hyperparameter Selection Across Adaptation Regimes.}
A limitation of our study is that hyperparameters are not tuned separately for each adaptation regime. Since the same optimization and method-specific settings are used across all trainable depth configurations, some of the observed ranking differences may reflect regime-dependent hyperparameter effects in addition to differences caused by trainable depth itself.

\paragraph{Narrow Axis of Regime Variation.}
A limitation of our study is that we vary the adaptation regime only through trainable depth. Although this is enough to reveal substantial changes in comparative method rankings, it covers only a narrow portion of the possible adaptation space. Alternative trainable subspace choices, such as sparse masks, low-rank updates, or structured pruning, may interact differently with the dynamics of continual learning. Therefore, our results should be interpreted as establishing sensitivity to trainable depth in particular, rather than as an exhaustive characterization of all forms of regime variation.

\paragraph{Restriction to task incremental continual learning.}
Our study is restricted to the task incremental setting, where task identity is available at both training and test time and inference is performed only over the logits of the current task. This setting allows for a controlled comparison of methods and regimes, but it is more structured than other continual learning scenarios. In particular, the role of the fine-tuning regime may differ in class-incremental or task-agnostic settings, where interference is typically stronger, and the absence of task labels makes the optimization problem more challenging.

\paragraph{Limited architectural scope.}
The empirical study is limited to a single backbone, namely ResNet-18. While this architecture provides a standard and well-understood testbed, different model families may exhibit different forms of regime sensitivity due to differences in depth, parameterization, feature reuse, or optimization dynamics. Extending the analysis to other architectures would help determine how broadly the observed effects transfer beyond the specific setting studied here.

\paragraph{Partial empirical validation of the optimization view.}
The projected optimization perspective introduced in Section~\ref{sec:our_approach} is intended to interpret regime sensitivity rather than to provide a complete mechanistic account. While the empirical results are consistent with this view, we do not directly measure all projected quantities across methods, tasks, and regimes. A more complete empirical validation of the proposed mechanism would require tracking these quantities more systematically throughout training.

\section{Conclusions}
\label{sec:conclusion}

We studied whether comparative conclusions in continual learning (CL) remain stable when the fine-tuning regime changes. Our main finding is that they often do not. When the trainable parameter subspace varies with trainable depth, the relative ranking of standard CL methods can change substantially, even when the architecture, task orders, and training pipeline are otherwise held fixed. This effect is consistent across multiple datasets and repeated task orderings, indicating that the adaptation regime is not a secondary implementation detail but a meaningful source of variation in CL evaluation.

Our analysis provides a simple lens for interpreting this result. By constraining optimization to a fixed trainable subspace, the fine-tuning regime changes the update directions available to both the task-fitting and preservation terms, thereby altering the effective optimization signal seen by the learner. Empirically, more permissive regimes are associated with larger update magnitudes, greater forgetting, and a stronger link between the two, which is consistent with the view that deeper adaptation exposes more interference-prone directions.

Taken together, these results suggest that CL benchmarks should treat the fine-tuning regime as an explicit experimental factor. Reporting results under a single adaptation setting can yield regime-specific comparative conclusions that do not necessarily generalize across training configurations. A more informative evaluation protocol should therefore account for the adaptation regime alongside algorithm choice and task order.

\section*{Future Work}
\label{sec: future_work} 
The limitations discussed in Section~\ref{sec:limitations} suggest several directions for future work. First, extending the empirical study beyond ResNet-18 to more recent convolutional and Transformer-based architectures would help establish whether the observed sensitivity of continual learning method rankings to adaptation regimes generalizes across different model families. Second, tuning optimization and method-specific hyperparameters independently for each adaptation regime would allow future studies to better disentangle the effects of the trainable parameter subspace from regime-dependent optimization choices. Third, the analysis could be broadened beyond task incremental learning to class-incremental, domain-incremental, and task-agnostic settings, where stronger interference and the absence of task labels may alter the interaction between adaptation regimes and forgetting. Finally, investigating richer forms of trainable subspaces-including parameter-efficient fine-tuning approaches such as low-rank updates, adapters, sparse masks, and structured pruning would provide a more comprehensive empirical validation of the proposed optimization perspective. Together, these extensions would test the breadth of the conclusions drawn here and clarify the extent to which regime sensitivity is a general property of continual learning evaluation.

\section*{Acknowledgment}
This research was supported by the project “Romanian Hub for Artificial Intelligence - HRIA”, Smart Growth, Digitization and Financial Instruments Program, 2021-2027, MySMIS no. 351416 and by the “European Lighthouse of AI for Sustainability - ELIAS”, Horizon Europe program (Grant No. 101120237).

\bibliographystyle{IEEEtran}
\bibliography{biblio}

\appendices

\section{\break Notation Summary}
We provide an overview of the mathematical notation in Table~\ref{tab:notation}.

\begin{table}[!h]
\centering
\small
\caption{Notation used in the theoretical analysis}
\begin{tabular}{p{0.20\linewidth} p{0.72\linewidth}}
\toprule
\textbf{Symbol} & \textbf{Meaning} \\
\midrule
$\mathcal{T}_1, \dots, \mathcal{T}_T$ & Sequence of tasks observed in continual learning. \\

$T$ & Number of tasks in the continual learning sequence. \\

$\theta \in \mathbb{R}^d$ & Full model parameter vector. \\

$d$ & Total number of model parameters. \\

$t$ & Task index. \\

$m$ & Optimization-step index within task $t$. \\

$\theta_t$ & Parameters after completing training on task $t$. \\

$\theta_t^{(m)}$ & Parameters after $m$ optimization steps on task $t$. \\

$\mathcal{L}_t(\theta)$ & Current-task loss at task $t$. \\

$\Omega_t(\theta;\theta_{1:t-1})$ & Method-specific term that preserves previously acquired knowledge. \\

$\lambda$ & Weight of the preservation term. \\

$\mathcal{J}_t(\theta)$ & Full task-$t$ objective, defined as $\mathcal{L}_t(\theta) + \lambda \Omega_t(\theta;\theta_{1:t-1})$. \\

$S \subseteq \{1,\dots,d\}$ & Fixed subset of trainable parameter coordinates defining an adaptation regime. \\

$P_S$ & Orthogonal projector that keeps coordinates in $S$ and zeros out all others. \\

$\eta$ & Step size used by the first-order optimizer. \\

$g_t^{(S)}(\theta)$ & Projected current-task update signal, defined as $P_S \nabla \mathcal{L}_t(\theta)$. \\

$r_t^{(S)}(\theta)$ & Projected preservation update signal, defined as $P_S \nabla \Omega_t(\theta;\theta_{1:t-1})$. \\

$\Gamma_t(S;\theta)$ & Interaction term between projected current-task and preservation signals, defined as $\langle g_t^{(S)}(\theta), r_t^{(S)}(\theta)\rangle$. \\

$P_S \nabla \mathcal{J}_t(\theta)$ & Projected gradient of the full task-$t$ objective. \\

$L$ & Smoothness constant of $\mathcal{J}_t$. \\

$\theta^+$ & One-step projected update from $\theta$, defined in \ref{eq:appendix_theta_plus}. \\

$\|\cdot\|$ & Euclidean norm. \\

$\langle \cdot,\cdot \rangle$ & Standard Euclidean inner product. \\
\bottomrule
\end{tabular}
\label{tab:notation}
\end{table}

\vspace{-3mm}

\section{\break Additional Theoretical Details}
\label{appendix:proofs}

For completeness, we restate the progress bound used in the main text.

\vspace{-4mm}

\paragraph{Progress bound under projected updates.}
Let
\begin{equation}
\theta^+ = \theta - \eta P_S \nabla \mathcal{J}_t(\theta),
\label{eq:appendix_theta_plus}
\end{equation}
where $P_S$ is the orthogonal projector associated with the fixed trainable subset $S$. Assume that $\mathcal{J}_t$ is $L$-smooth on a neighborhood containing both $\theta$ and $\theta^+$. Then
\begin{equation}
\mathcal{J}_t(\theta^+)
\le
\mathcal{J}_t(\theta)
-
\eta \|P_S \nabla \mathcal{J}_t(\theta)\|^2
+
\frac{L\eta^2}{2}\|P_S \nabla \mathcal{J}_t(\theta)\|^2.
\label{eq:appendix_descent_1}
\end{equation}
Equivalently,
\begin{equation}
\mathcal{J}_t(\theta^+)
\le
\mathcal{J}_t(\theta)
-
\eta \Bigl(1-\frac{L\eta}{2}\Bigr)\|P_S \nabla \mathcal{J}_t(\theta)\|^2.
\label{eq:appendix_descent_2}
\end{equation}
In particular, if $0 < \eta \le 1/L$, then
\begin{equation}
\mathcal{J}_t(\theta^+)
\le
\mathcal{J}_t(\theta)
-
\frac{\eta}{2}\|P_S \nabla \mathcal{J}_t(\theta)\|^2.
\label{eq:appendix_descent_3}
\end{equation}

\paragraph{Proof.}
Since $\mathcal{J}_t$ is $L$-smooth, for any $x,y$ in the neighborhood of interest,
\begin{equation}
\mathcal{J}_t(y)
\le
\mathcal{J}_t(x)
+
\langle \nabla \mathcal{J}_t(x), y-x \rangle
+
\frac{L}{2}\|y-x\|^2.
\label{eq:appendix_smoothness}
\end{equation}
Apply \ref{eq:appendix_smoothness} with $x=\theta$ and $y=\theta^+$. Using \ref{eq:appendix_theta_plus},
\begin{equation}
\theta^+ - \theta = -\eta P_S \nabla \mathcal{J}_t(\theta).
\label{eq:appendix_step_diff}
\end{equation}
Substituting into \ref{eq:appendix_smoothness} gives
\begin{align}
\mathcal{J}_t(\theta^+)
&\le
\mathcal{J}_t(\theta)
+
\left\langle \nabla \mathcal{J}_t(\theta), \theta^+ - \theta \right\rangle
+
\frac{L}{2}\|\theta^+ - \theta\|^2 \\
&=
\mathcal{J}_t(\theta)
-
\eta \left\langle \nabla \mathcal{J}_t(\theta), P_S \nabla \mathcal{J}_t(\theta) \right\rangle
+
\frac{L\eta^2}{2}\|P_S \nabla \mathcal{J}_t(\theta)\|^2
\label{eq:appendix_before_projector_identity}
\end{align}

Now, $P_S$ is an orthogonal projector. Hence, it is symmetric and idempotent:
\begin{equation}
P_S^\top = P_S,
\qquad
P_S^2 = P_S.
\label{eq:appendix_projector_properties}
\end{equation}
Therefore,
\begin{align}
\left\langle \nabla \mathcal{J}_t(\theta), P_S \nabla \mathcal{J}_t(\theta) \right\rangle
&=
\nabla \mathcal{J}_t(\theta)^\top P_S \nabla \mathcal{J}_t(\theta) \\
&=
\nabla \mathcal{J}_t(\theta)^\top P_S^\top P_S \nabla \mathcal{J}_t(\theta) \\
&=
\|P_S \nabla \mathcal{J}_t(\theta)\|^2.
\label{eq:appendix_projector_identity}
\end{align}
Substituting \ref{eq:appendix_projector_identity} into \ref{eq:appendix_before_projector_identity} yields \ref{eq:appendix_descent_1}. Rearranging gives \ref{eq:appendix_descent_2}. Finally, if $0<\eta\le 1/L$, then
\begin{equation}
1-\frac{L\eta}{2}\ge \frac{1}{2},
\end{equation}
which implies \ref{eq:appendix_descent_3}. \hfill $\square$

\section{\break Additional experimental values}
\label{additional_exp}

Tables~\ref{tab:cifar100}, \ref{tab:mnist}, \ref{tab:fashion}, \ref{tab:qmnist}, and \ref{tab:kmnist} report the numerical values underlying the results shown in Figures \ref{fig:main_results_fashion_mnist} and \ref{fig:main_results_all_datasets}, listing average accuracy and average forgetting per training regime for each algorithm across CIFAR-100, MNIST, Fashion-MNIST, QMNIST, and KMNIST, together with standard deviations computed over task orderings.

\begin{table*}[t]
\centering
\caption{Average Accuracy and Average Forgetting per ordering $\pm$ Standard deviation across different training regimes on the CIFAR-100 dataset}
\resizebox{\textwidth}{!}{
\begin{tabular}{lcc cc cc cc cc}
\toprule
\multirow{2}{*}{\textbf{Algorithm / Regime}} 
& \multicolumn{2}{c}{\textbf{Last Block}} 
& \multicolumn{2}{c}{\textbf{Last 2 Blocks}} 
& \multicolumn{2}{c}{\textbf{Last 3 Blocks}} 
& \multicolumn{2}{c}{\textbf{Last 6 Blocks}} 
& \multicolumn{2}{c}{\textbf{Full Finetune (8 Blocks)}} \\
\cmidrule(lr){2-3} \cmidrule(lr){4-5} \cmidrule(lr){6-7} \cmidrule(lr){8-9} \cmidrule(lr){10-11}
& Acc & Forget 
& Acc & Forget 
& Acc & Forget 
& Acc & Forget 
& Acc & Forget \\
\midrule

online EWC 
& $0.52 \pm 0.009$ & $0.02 \pm 0.008$ 
& $0.59 \pm 0.010$ & $0.02 \pm 0.006$ 
& $0.62 \pm 0.013$ & $0.02 \pm 0.009$ 
& $0.64 \pm 0.017$ & $0.03 \pm 0.006$ 
& $0.65 \pm 0.016$ & $0.03 \pm 0.008$ \\

LwF 
& $0.62 \pm 0.008$ & $0.05 \pm 0.005$ 
& $0.70 \pm 0.006$ & $0.08 \pm 0.004$ 
& $0.75 \pm 0.004$ & $0.08 \pm 0.005$ 
& $0.79 \pm 0.008$ & $0.07 \pm 0.007$ 
& $0.80 \pm 0.008$ & $0.07 \pm 0.008$ \\

SI 
& $0.57 \pm 0.007$ & $0.04 \pm 0.006$ 
& $0.61 \pm 0.013$ & $0.09 \pm 0.012$ 
& $0.62 \pm 0.021$ & $0.14 \pm 0.019$ 
& $0.62 \pm 0.018$ & $0.18 \pm 0.015$ 
& $0.63 \pm 0.017$ & $0.18 \pm 0.017$ \\

GEM 
& $0.52 \pm 0.008$ & $0.15 \pm 0.010$ 
& $0.55 \pm 0.010$ & $0.21 \pm 0.011$ 
& $0.57 \pm 0.012$ & $0.26 \pm 0.011$ 
& $0.61 \pm 0.012$ & $0.26 \pm 0.014$ 
& $0.71 \pm 0.008$ & $0.16 \pm 0.011$ \\

DER
& $0.63 \pm 0.007$ & $0.02 \pm 0.006$ 
& $0.71 \pm 0.003$ & $0.03 \pm 0.006$ 
& $0.75 \pm 0.007$ & $0.03 \pm 0.006$ 
& $0.79 \pm 0.006$ & $0.03 \pm 0.007$ 
& $0.80 \pm 0.003$ & $0.03 \pm 0.004$ \\

\bottomrule
\end{tabular}
}
\label{tab:cifar100}
\end{table*}

\begin{table*}[t]
\centering
\caption{Average Accuracy and Average Forgetting per ordering $\pm$ Standard deviation across different training regimes on the MNIST dataset}
\resizebox{\textwidth}{!}{
\begin{tabular}{lcc cc cc cc cc}
\toprule
\multirow{2}{*}{\textbf{Algorithm / Regime}} 
& \multicolumn{2}{c}{\textbf{Last Block}} 
& \multicolumn{2}{c}{\textbf{Last 2 Blocks}} 
& \multicolumn{2}{c}{\textbf{Last 3 Blocks}} 
& \multicolumn{2}{c}{\textbf{Last 6 Blocks}} 
& \multicolumn{2}{c}{\textbf{Full Finetune (8 Blocks)}} \\
\cmidrule(lr){2-3} \cmidrule(lr){4-5} \cmidrule(lr){6-7} \cmidrule(lr){8-9} \cmidrule(lr){10-11}
& Acc & Forget 
& Acc & Forget 
& Acc & Forget 
& Acc & Forget 
& Acc & Forget \\
\midrule

online EWC 
& $0.98 \pm 0.012$ & $0.02 \pm 0.015$ 
& $0.91 \pm 0.082$ & $0.11 \pm 0.103$ 
& $0.87 \pm 0.083$ & $0.16 \pm 0.104$ 
& $0.86 \pm 0.060$ & $0.18 \pm 0.075$ 
& $0.89 \pm 0.078$ & $0.14 \pm 0.097$ \\

LwF 
& $0.99 \pm 0.002$ & $0.01 \pm 0.003$ 
& $0.99 \pm 0.007$ & $0.01 \pm 0.009$ 
& $0.99 \pm 0.007$ & $0.01 \pm 0.009$ 
& $0.99 \pm 0.004$ & $0.01 \pm 0.005$ 
& $0.99 \pm 0.011$ & $0.01 \pm 0.014$ \\

SI 
& $0.96 \pm 0.018$ & $0.04 \pm 0.021$ 
& $0.86 \pm 0.090$ & $0.17 \pm 0.113$ 
& $0.84 \pm 0.095$ & $0.20 \pm 0.119$ 
& $0.80 \pm 0.072$ & $0.25 \pm 0.090$ 
& $0.80 \pm 0.056$ & $0.25 \pm 0.070$ \\

GEM 
& $0.96 \pm 0.024$ & $0.04 \pm 0.030$ 
& $0.94 \pm 0.038$ & $0.08 \pm 0.047$ 
& $0.92 \pm 0.059$ & $0.10 \pm 0.075$ 
& $0.93 \pm 0.053$ & $0.09 \pm 0.067$ 
& $0.95 \pm 0.037$ & $0.06 \pm 0.047$ \\

DER
& $0.99 \pm 0.004$ & $0.01 \pm 0.005$ 
& $1.00 \pm 0.002$ & $0.00 \pm 0.003$ 
& $1.00 \pm 0.002$ & $0.00 \pm 0.002$ 
& $1.00 \pm 0.003$ & $0.00 \pm 0.003$ 
& $1.00 \pm 0.004$ & $0.00 \pm 0.004$ \\

\bottomrule
\end{tabular}
}
\label{tab:mnist}
\end{table*}

\begin{table*}[t]
\centering
\caption{Average Accuracy and Average Forgetting per ordering $\pm$ Standard deviation across different training regimes on the Fashion MNIST dataset}
\resizebox{\textwidth}{!}{
\begin{tabular}{lcc cc cc cc cc}
\toprule
\multirow{2}{*}{\textbf{Algorithm / Regime}} 
& \multicolumn{2}{c}{\textbf{Last Block}} 
& \multicolumn{2}{c}{\textbf{Last 2 Blocks}} 
& \multicolumn{2}{c}{\textbf{Last 3 Blocks}} 
& \multicolumn{2}{c}{\textbf{Last 6 Blocks}} 
& \multicolumn{2}{c}{\textbf{Full Finetune (8 Blocks)}} \\
\cmidrule(lr){2-3} \cmidrule(lr){4-5} \cmidrule(lr){6-7} \cmidrule(lr){8-9} \cmidrule(lr){10-11}
& Acc & Forget 
& Acc & Forget 
& Acc & Forget 
& Acc & Forget 
& Acc & Forget \\
\midrule

online EWC 
& $0.92 \pm 0.043$ & $0.09 \pm 0.054$ 
& $0.91 \pm 0.057$ & $0.10 \pm 0.072$ 
& $0.86 \pm 0.063$ & $0.16 \pm 0.079$ 
& $0.81 \pm 0.082$ & $0.23 \pm 0.103$ 
& $0.85 \pm 0.041$ & $0.17 \pm 0.051$ \\

LwF 
& $0.91 \pm 0.042$ & $0.10 \pm 0.053$ 
& $0.90 \pm 0.059$ & $0.12 \pm 0.074$ 
& $0.86 \pm 0.084$ & $0.17 \pm 0.105$ 
& $0.85 \pm 0.077$ & $0.17 \pm 0.097$ 
& $0.82 \pm 0.049$ & $0.21 \pm 0.061$ \\

SI 
& $0.92 \pm 0.027$ & $0.09 \pm 0.035$ 
& $0.93 \pm 0.022$ & $0.08 \pm 0.027$ 
& $0.88 \pm 0.057$ & $0.14 \pm 0.071$ 
& $0.89 \pm 0.059$ & $0.13 \pm 0.075$ 
& $0.87 \pm 0.063$ & $0.16 \pm 0.080$ \\

GEM 
& $0.97 \pm 0.012$ & $0.02 \pm 0.013$ 
& $0.96 \pm 0.016$ & $0.03 \pm 0.019$ 
& $0.95 \pm 0.030$ & $0.05 \pm 0.041$ 
& $0.91 \pm 0.037$ & $0.08 \pm 0.051$ 
& $0.94 \pm 0.034$ & $0.06 \pm 0.042$ \\

DER
& $0.96 \pm 0.013$ & $0.04 \pm 0.016$ 
& $0.97 \pm 0.015$ & $0.03 \pm 0.019$ 
& $0.95 \pm 0.018$ & $0.05 \pm 0.022$ 
& $0.96 \pm 0.012$ & $0.05 \pm 0.015$ 
& $0.94 \pm 0.017$ & $0.06 \pm 0.021$ \\

\bottomrule
\end{tabular}
}
\label{tab:fashion}
\end{table*}

\begin{table*}[t]
\centering
\caption{Average Accuracy and Average Forgetting per ordering $\pm$ Standard deviation across different training regimes on the QMNIST dataset}
\resizebox{\textwidth}{!}{
\begin{tabular}{lcc cc cc cc cc}
\toprule
\multirow{2}{*}{\textbf{Algorithm / Regime}} 
& \multicolumn{2}{c}{\textbf{Last Block}} 
& \multicolumn{2}{c}{\textbf{Last 2 Blocks}} 
& \multicolumn{2}{c}{\textbf{Last 3 Blocks}} 
& \multicolumn{2}{c}{\textbf{Last 6 Blocks}} 
& \multicolumn{2}{c}{\textbf{Full Finetune (8 Blocks)}} \\
\cmidrule(lr){2-3} \cmidrule(lr){4-5} \cmidrule(lr){6-7} \cmidrule(lr){8-9} \cmidrule(lr){10-11}
& Acc & Forget 
& Acc & Forget 
& Acc & Forget 
& Acc & Forget 
& Acc & Forget \\
\midrule

online EWC 
& $0.98 \pm 0.008$ & $0.01 \pm 0.010$ 
& $0.91 \pm 0.073$ & $0.10 \pm 0.090$ 
& $0.90 \pm 0.061$ & $0.12 \pm 0.076$ 
& $0.92 \pm 0.053$ & $0.10 \pm 0.067$ 
& $0.85 \pm 0.084$ & $0.19 \pm 0.104$ \\

LwF 
& $0.99 \pm 0.004$ & $0.01 \pm 0.005$ 
& $0.99 \pm 0.004$ & $0.01 \pm 0.004$ 
& $0.99 \pm 0.004$ & $0.01 \pm 0.005$ 
& $0.98 \pm 0.030$ & $0.03 \pm 0.036$ 
& $0.98 \pm 0.026$ & $0.03 \pm 0.033$ \\

SI 
& $0.97 \pm 0.025$ & $0.03 \pm 0.033$ 
& $0.85 \pm 0.095$ & $0.18 \pm 0.120$ 
& $0.79 \pm 0.127$ & $0.25 \pm 0.158$ 
& $0.82 \pm 0.068$ & $0.23 \pm 0.085$ 
& $0.85 \pm 0.101$ & $0.18 \pm 0.127$ \\

GEM 
& $0.96 \pm 0.014$ & $0.04 \pm 0.018$ 
& $0.93 \pm 0.047$ & $0.08 \pm 0.059$ 
& $0.93 \pm 0.045$ & $0.08 \pm 0.056$ 
& $0.94 \pm 0.052$ & $0.08 \pm 0.065$ 
& $0.95 \pm 0.049$ & $0.06 \pm 0.061$ \\

DER 
& $0.99 \pm 0.004$ & $0.01 \pm 0.004$ 
& $0.99 \pm 0.003$ & $0.00 \pm 0.003$ 
& $1.00 \pm 0.002$ & $0.00 \pm 0.002$ 
& $0.99 \pm 0.004$ & $0.00 \pm 0.005$ 
& $0.99 \pm 0.008$ & $0.01 \pm 0.010$ \\

\bottomrule
\end{tabular}
}
\label{tab:qmnist}
\end{table*}

\begin{table*}[t]
\caption{Average Accuracy and Average Forgetting per ordering $\pm$ Standard deviation across different training regimes on the KMNIST dataset}
\centering
\resizebox{\textwidth}{!}{
\begin{tabular}{lcc cc cc cc cc}
\toprule
\multirow{2}{*}{\textbf{Algorithm / Regime}} 
& \multicolumn{2}{c}{\textbf{Last Block}} 
& \multicolumn{2}{c}{\textbf{Last 2 Blocks}} 
& \multicolumn{2}{c}{\textbf{Last 3 Blocks}} 
& \multicolumn{2}{c}{\textbf{Last 6 Blocks}} 
& \multicolumn{2}{c}{\textbf{Full Finetune (8 Blocks)}} \\
\cmidrule(lr){2-3} \cmidrule(lr){4-5} \cmidrule(lr){6-7} \cmidrule(lr){8-9} \cmidrule(lr){10-11}
& Acc & Forget 
& Acc & Forget 
& Acc & Forget 
& Acc & Forget 
& Acc & Forget \\
\midrule

online EWC 
& $0.94 \pm 0.010$ & $-0.00 \pm 0.009$ 
& $0.95 \pm 0.016$ & $0.03 \pm 0.022$ 
& $0.93 \pm 0.028$ & $0.06 \pm 0.037$ 
& $0.92 \pm 0.033$ & $0.08 \pm 0.045$ 
& $0.93 \pm 0.026$ & $0.06 \pm 0.033$ \\

LwF 
& $0.95 \pm 0.009$ & $0.02 \pm 0.008$ 
& $0.97 \pm 0.012$ & $0.02 \pm 0.014$ 
& $0.97 \pm 0.008$ & $0.03 \pm 0.010$ 
& $0.96 \pm 0.008$ & $0.03 \pm 0.010$ 
& $0.97 \pm 0.009$ & $0.03 \pm 0.010$ \\

SI 
& $0.91 \pm 0.025$ & $0.05 \pm 0.037$ 
& $0.86 \pm 0.035$ & $0.14 \pm 0.045$ 
& $0.86 \pm 0.050$ & $0.15 \pm 0.061$ 
& $0.82 \pm 0.047$ & $0.20 \pm 0.057$ 
& $0.84 \pm 0.081$ & $0.17 \pm 0.099$ \\

GEM 
& $0.91 \pm 0.014$ & $0.07 \pm 0.017$ 
& $0.91 \pm 0.016$ & $0.10 \pm 0.019$ 
& $0.91 \pm 0.016$ & $0.10 \pm 0.021$ 
& $0.92 \pm 0.014$ & $0.09 \pm 0.018$ 
& $0.92 \pm 0.020$ & $0.09 \pm 0.026$ \\

DER 
& $0.95 \pm 0.011$ & $0.00 \pm 0.008$ 
& $0.98 \pm 0.003$ & $0.01 \pm 0.005$ 
& $0.98 \pm 0.006$ & $0.01 \pm 0.007$ 
& $0.98 \pm 0.004$ & $0.01 \pm 0.006$ 
& $0.98 \pm 0.006$ & $0.01 \pm 0.008$ \\

\bottomrule
\end{tabular}
}
\label{tab:kmnist}
\end{table*}

\vspace{5mm}

\begin{IEEEbiography}[{\includegraphics[width=1in,height=1.25in,clip,keepaspectratio]{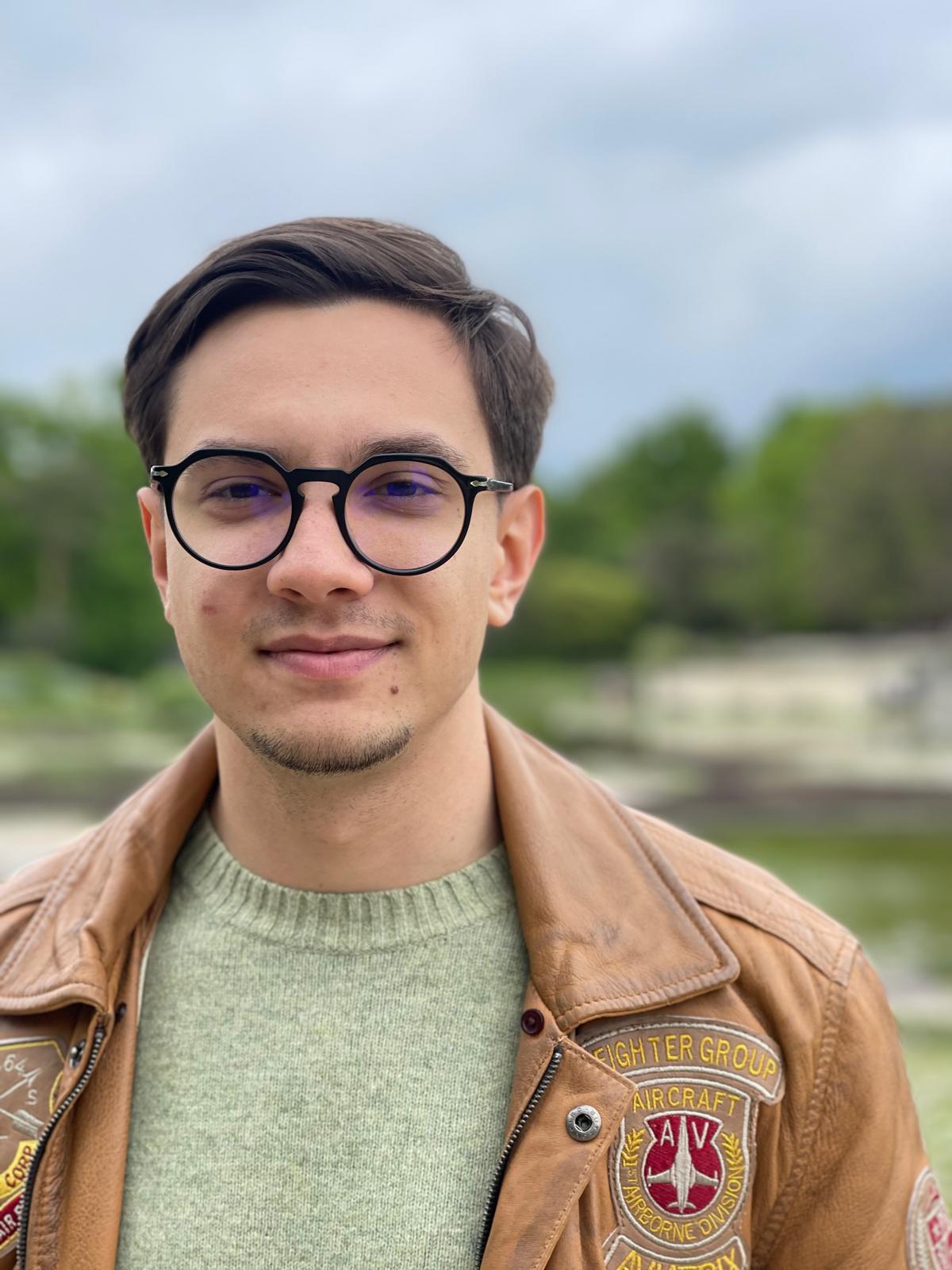}}]{Paul-Tiberiu Iordache} received the B.Sc.
degree in computer science (minor in mechanical
engineering) and the M.Sc. degree in artificial
intelligence (algorithms, artificial intelligence,
interactions and decision) from Sorbonne University, Paris, France, in 2023 and 2025, respectively.
He is currently a Machine Learning Researcher
with Bitdefender, Bucharest, Romania, where he
works on continual learning and develops anomaly
detection systems for networks traffic using timeseries foundation models. Previously, he worked at CEA, Paris, France,
where he applied deep reinforcement learning to moving target defense
strategies for microservice architectures, work that resulted in a paper
accepted at the UCC Conference. He also worked on multi-agent onboarding
systems, facial recognition, age-classification models, and satellite-imagery
vehicle detection. His research interests include reinforcement learning,
continual learning, computer vision, deep learning, time-series modeling,
anomaly detection, and machine learning applications in cybersecurity.
\end{IEEEbiography}

\begin{IEEEbiography}[{\includegraphics[width=1in,height=1.25in,clip,keepaspectratio]{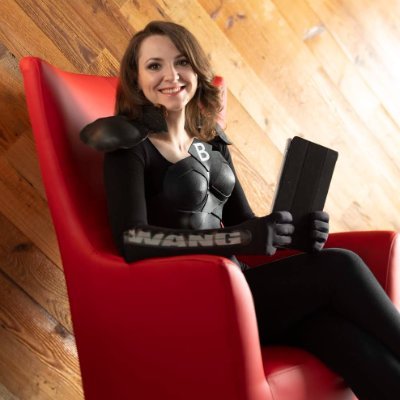}}]{Elena BURCEANU} is an AI Research Lead at Bitdefender, where she leads research in machine learning and its applications to cybersecurity. Her research focuses on robust and trustworthy AI, with particular interests in generalization under distribution shifts, spurious correlations, novelty detection, continual learning, and the reliability of foundation models. She has published research at leading AI and computer vision venues, including ICLR, NeurIPS, ICCV, ECCV, IJCAI, EMNLP, NAACL, BMVC, WACV, and Artificial Intelligence, spanning topics from unsupervised video understanding and object tracking to anomaly detection, authorship verification, robust novelty detection, and bias and spurious correlations in foundation models. Elena received her PhD in Computer Science from the University of Bucharest in 2022, following BSc and MSc degrees in Computer Science and Distributed Systems from University Politehnica of Bucharest. She is actively involved in the AI research community and has served as Associate Program Chair for NeurIPS 2025. She is also a co-inventor on eight granted patents and is the Local Chair of CoLLAs 2026 in Bucharest.
\end{IEEEbiography}

\begin{IEEEbiography}[{\includegraphics[width=1in,height=1.25in,clip,keepaspectratio]{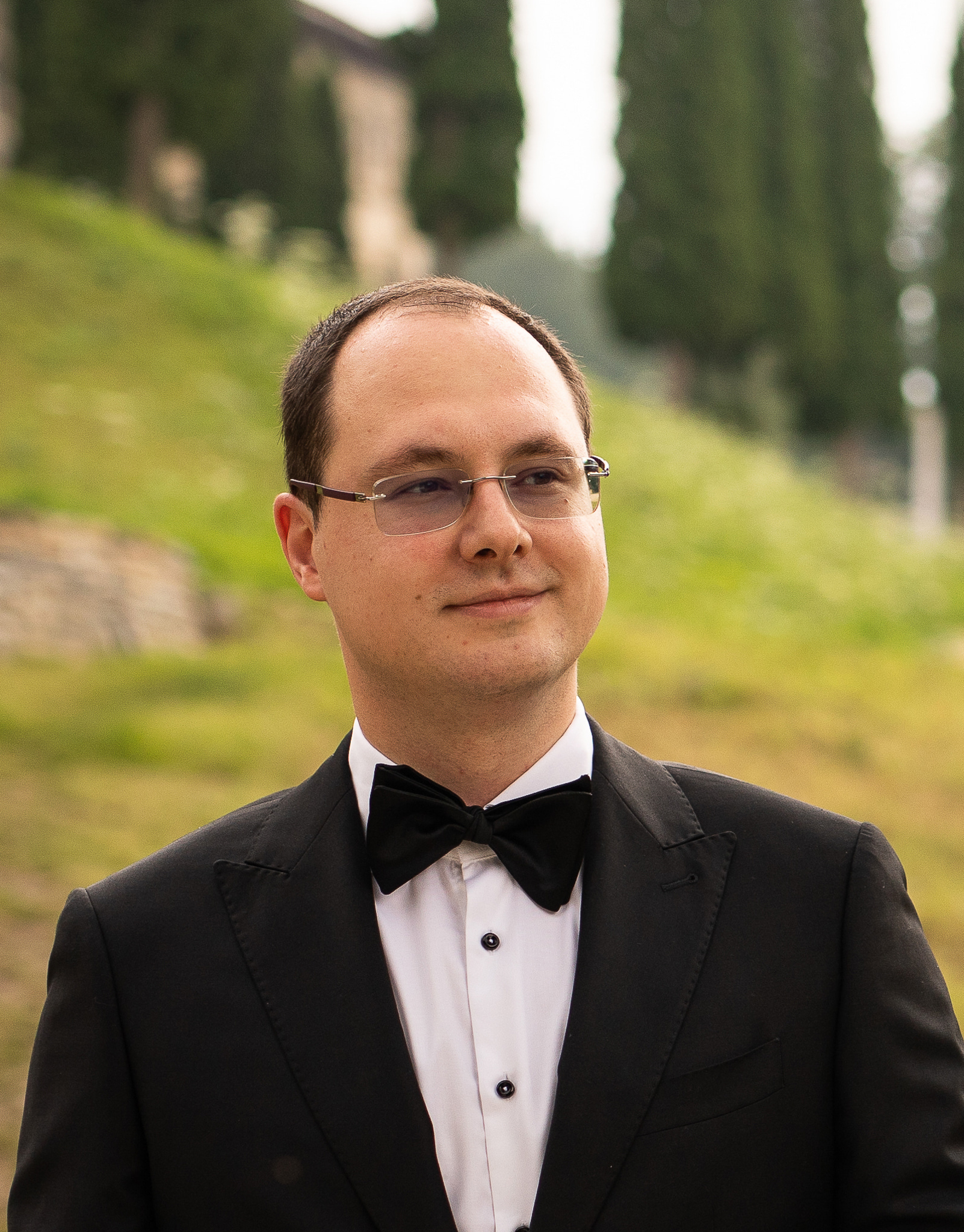}}]{Mihai Dascalu} is a Full Professor of Computer Science at the National University of Science and Technology Politehnica Bucharest, where he teaches and coordinates several core courses, including Object-Oriented Programming, Algorithm Design, and Introduction to Machine Learning. He holds a double Ph.D. in Computer Science (UPB) and Educational Sciences (University Grenoble Alpes, France), both awarded with the highest distinctions. His research focuses on Natural Language Processing, AI in education, and computational modeling of discourse. He has participated in and led numerous national and international research projects, in collaboration with academic institutions, government agencies, and industry partners. He has authored more than 300 publications, including over 40 articles at top-tier international conferences and more than 30 Q1/Q2 journal papers.
\end{IEEEbiography}

\EOD

\end{document}